%% file: neurips_2025.tex
\documentclass{article}

\PassOptionsToPackage{numbers, square, sort&compress}{natbib}
    \usepackage[final]{neurips_2025}

\usepackage[utf8]{inputenc} % allow utf-8 input
\usepackage[T1]{fontenc}    % use 8-bit T1 fonts
\usepackage{hyperref}       % hyperlinks
\usepackage{url}            % simple URL typesetting
\usepackage{booktabs}       % professional-quality tables
\usepackage{amsfonts}       % blackboard math symbols
\usepackage{nicefrac}       % compact symbols for 1/2, etc.
\usepackage{microtype}      % microtypography
\usepackage{xcolor}         % colors
\usepackage{amsmath}
\usepackage{enumitem}
\usepackage{mathtools}
\usepackage{tikz}
\usetikzlibrary{decorations.pathreplacing,positioning}
\usepackage{amsthm}
\usepackage{multirow}
\usepackage{makecell}
\usepackage{subcaption}
\usepackage{booktabs}
\usepackage{pifont}
\usepackage{wrapfig}
\usepackage{placeins}

\newcommand{\vb}[1]{\mathbf{#1}}

\newtheorem{theorem}{Theorem}[section]
\newtheorem{corollary}{Corollary}[section]
\newtheorem{remark}{Remark}[section]
\newtheorem{proposition}{Proposition}[section]
\newtheorem{definition}{Definition}[section]

\title{A Pre-training Framework for Relational Data with Information-theoretic Principles}

\author{%
Quang Truong$^1$\,\,,  
Zhikai Chen$^1$\,\,, 
Mingxuan Ju$^2$\,\,, 
Tong Zhao$^2$\,\,, 
Neil Shah$^2$\,\,, 
Jiliang Tang$^1$ \\
 $^1$Michigan State University,
 $^2$Snap Inc.\\
 \texttt{\{truongc4, chenzh85,tangjili\}@msu.edu}
 \\
 \texttt{\{mju, tong, nshah\}@snap.com}
}

\begin{document}

\maketitle

\begin{abstract}

% \czk{Here we have a jump from relational database to RDL, RDL is not defined}

% \czk{A general concern for the motivation part of abstract and introduction is that it's probably not a good idea to motivate from that SSL can not work. It's generally not surprising to see SSL for CV can not be directly adapted for tabular data. If we start from the point that the previous method cannot work, then we also need to discuss why methods like SSL for tabular data \url{https://arxiv.org/pdf/2405.07414} are not applicable in our case. Maybe discussed later.}

Relational databases underpin critical infrastructure across a wide range of domains, yet the design of generalizable pre-training strategies for learning from relational databases remains an open challenge due to task heterogeneity. Specifically, there exist many possible downstream tasks, as tasks are defined based on relational schema graphs, temporal dependencies, and SQL-defined label logics. An effective pre-training framework is desired to take these factors into account in order to obtain task-aware representations. By incorporating knowledge of the underlying distribution that drives label generation, downstream tasks can benefit from relevant side-channel information. To bridge this gap, we introduce Task Vector Estimation (TVE), a novel pre-training framework that constructs predictive supervisory signals via set-based aggregation over schema traversal graphs, explicitly modeling next-window relational dynamics. We formalize our approach through an information-theoretic lens, demonstrating that task-informed representations retain more relevant signals than those obtained without task priors. Extensive experiments on the RelBench benchmark show that TVE consistently outperforms traditional pre-training baselines. Our findings advocate for pre-training objectives that encode task heterogeneity and temporal structure as design principles for predictive modeling on relational databases. Our code is publicly available at \href{https://github.com/quang-truong/task-vector-estimation}{\texttt{https://github.com/quang-truong/task-vector-estimation}}.

\end{abstract}

\input{sections/1_Introduction}

\input{sections/2_Preliminaries}
\input{sections/3_Pretraining_Framework}
\input{sections/4_Task_Vector_Estimation}
\input{sections/5_Experiments}
\input{sections/6_Discussion_and_Related_Works}
% \input{sections/7_Conclusion}
\section{Acknowledgements}
Quang Truong, Zhikai Chen, and Jiliang Tang are supported by the National Science Foundation (NSF) under grant numbers CNS2321416, IIS2212032, IIS2212144, IIS 2504089, DUE2234015, CNS2246050, DRL2405483 and IOS2035472, the Michigan Department of Agriculture and Rural Development, US Dept of Commerce, Gates Foundation, Amazon Faculty Award, Meta, NVIDIA, Microsoft and SNAP.

\bibliographystyle{plainnat}
\bibliography{references}

\clearpage

\newpage
\section*{NeurIPS Paper Checklist}

\begin{enumerate}

\item {\bf Claims}
    \item[] Question: Do the main claims made in the abstract and introduction accurately reflect the paper's contributions and scope?
    \item[] Answer: \answerYes{}
    \item[] Justification: The abstract and introduce accurately describe the proposed framework, namely Task Vector Estimation and theoretical contributions in Section~\ref{sec:predictive}. Experimental results with performance gains are included in Section ~\ref{sec:experiments}.

\item {\bf Limitations}
    \item[] Question: Does the paper discuss the limitations of the work performed by the authors?
    \item[] Answer: \answerYes
    \item[] Justification: Limitations are discussed in Section~ \ref{sec:related}.

\item {\bf Theory assumptions and proofs}
    \item[] Question: For each theoretical result, does the paper provide the full set of assumptions and a complete (and correct) proof?
    \item[] Answer: \answerYes{}
    \item[] Justification: The theoretical assumptions are clearly stated in the statements in main texts and Appendix \ref{appendix:proof}.

    \item {\bf Experimental result reproducibility}
    \item[] Question: Does the paper fully disclose all the information needed to reproduce the main experimental results of the paper to the extent that it affects the main claims and/or conclusions of the paper (regardless of whether the code and data are provided or not)?
    \item[] Answer: \answerYes{}
    \item[] Justification: The paper discusses the proposed method in depth in Section \ref{sec:overview} and Section \ref{sec:predictive}, with experiment details in Section \ref{sec:experiments} and Appendix \ref{appendix:experiments}.

\item {\bf Open access to data and code}
    \item[] Question: Does the paper provide open access to the data and code, with sufficient instructions to faithfully reproduce the main experimental results, as described in supplemental material?
    \item[] Answer: \answerYes{} % Replace by \answerYes{}, \answerNo{}, or \answerNA{}.
    \item[] Justification: Codes, with sufficient instructions, are included in a public URL.

\item {\bf Experimental setting/details}
    \item[] Question: Does the paper specify all the training and test details (e.g., data splits, hyperparameters, how they were chosen, type of optimizer, etc.) necessary to understand the results?
    \item[] Answer: \answerYes{} % Replace by \answerYes{}, \answerNo{}, or \answerNA{}.
    \item[] Justification: Detailed experimental settings are stated in both main texts and Appendix.

\item {\bf Experiment statistical significance}
    \item[] Question: Does the paper report error bars suitably and correctly defined or other appropriate information about the statistical significance of the experiments?
    \item[] Answer: \answerYes{} % Replace by \answerYes{}, \answerNo{}, or \answerNA{}.
    \item[] Justification: Result tables include mean performance and standard deviation for every experiment.

\item {\bf Experiments compute resources}
    \item[] Question: For each experiment, does the paper provide sufficient information on the computer resources (type of compute workers, memory, time of execution) needed to reproduce the experiments?
    \item[] Answer: \answerYes{}
    \item[] Justification: Computational details are included in Appendix, where we state the CPU and GPU of machines we use.
    
\item {\bf Code of ethics}
    \item[] Question: Does the research conducted in the paper conform, in every respect, with the NeurIPS Code of Ethics \url{https://neurips.cc/public/EthicsGuidelines}?
    \item[] Answer: \answerYes{} % Replace by \answerYes{}, \answerNo{}, or \answerNA{}.
    \item[] Justification:

\item {\bf Broader impacts}
    \item[] Question: Does the paper discuss both potential positive societal impacts and negative societal impacts of the work performed?
    \item[] Answer: \answerNA
    \item[] Justification: We do not observe any negative societal impacts for our research.
    
\item {\bf Safeguards}
    \item[] Question: Does the paper describe safeguards that have been put in place for responsible release of data or models that have a high risk for misuse (e.g., pretrained language models, image generators, or scraped datasets)?
    \item[] Answer: \answerNA{}
    \item[] Justification: We do not observe any risks associated with our research.

\item {\bf Licenses for existing assets}
    \item[] Question: Are the creators or original owners of assets (e.g., code, data, models), used in the paper, properly credited and are the license and terms of use explicitly mentioned and properly respected?
    \item[] Answer: \answerYes{}
    \item[] Justification: We have given proper attributions to the original authors cited in the paper.

\item {\bf New assets}
    \item[] Question: Are new assets introduced in the paper well documented and is the documentation provided alongside the assets?
    \item[] Answer: \answerYes{}
    \item[] Justification: We provide details of our proposed tasks in the appendix and codes in the public URL.

\item {\bf Crowdsourcing and research with human subjects}
    \item[] Question: For crowdsourcing experiments and research with human subjects, does the paper include the full text of instructions given to participants and screenshots, if applicable, as well as details about compensation (if any)? 
    \item[] Answer: \answerNA{}
    \item[] Justification: Our paper does not involve crowdsourcing nor research with human subjects.

\item {\bf Institutional review board (IRB) approvals or equivalent for research with human subjects}
    \item[] Question: Does the paper describe potential risks incurred by study participants, whether such risks were disclosed to the subjects, and whether Institutional Review Board (IRB) approvals (or an equivalent approval/review based on the requirements of your country or institution) were obtained?
    \item[] Answer: \answerNA{}
    \item[] Justification: Our paper does not involve crowdsourcing nor research with human subjects.

\item {\bf Declaration of LLM usage}
    \item[] Question: Does the paper describe the usage of LLMs if it is an important, original, or non-standard component of the core methods in this research? Note that if the LLM is used only for writing, editing, or formatting purposes and does not impact the core methodology, scientific rigorousness, or originality of the research, declaration is not required.
    %this research? 
    \item[] Answer: \answerNA{}
    \item[] Justification: We only use LLMs for editing writing.

\end{enumerate}

\input{sections/0_Appendix}

\end{document}

%% file: sections/1_Introduction.tex
\section{Introduction}

Relational databases (RDBs) have long served as the de facto standard for managing data storage across diverse industries~\cite{halpin2010information, aditya2002banks, agrawal2001storage}, and learning tasks over RDBs depend on domain expertise for manually engineering meaningful features \cite{fey2023relationaldeeplearninggraph, kanter2015deep}.
% While manual feature engineering can yield effective models, \jt{the human efforts are not a concern for feature engineering??} it inherently introduces bias, susceptibility to human errors, and the risk of overlooking valuable predictive signals.
While manual feature engineering can yield effective models, it demands significant human time and expertise, and it inherently introduces bias, susceptibility to human errors, and the risk of overlooking valuable predictive signals.
Relational Deep Learning (RDL) \cite{fey2023relationaldeeplearninggraph, cvitkovic2020supervisedlearningrelationaldatabases} has emerged as a powerful paradigm that exploits relational structures---defined by primary-foreign key (PK-FK) links across tables. By treating RDBs as graphs of tables, we can leverage existing state-of-the-art tabular feature encoders \cite{gorishniy2021revisitingdltabular, chen2023trompt} and Graph Neural Networks (GNNs) \cite{kipf2017semisupervised, hamilton2017inductive, gilmer2017messagepassing} to extract meaningful embeddings in an end-to-end manner.
% Specifically, rows within tables are modeled as nodes, while primary-foreign key relationships form edges in a temporal heterogeneous graph. The resulting multi-table relational data is then encoded using tabular feature encoders \cite{tabular_encoder} and processed by graph neural networks (GNNs) \cite{gnn_reference} to extract meaningful embeddings for predictive modeling. 
This integrated approach has already demonstrated significant improvements in efficiency over conventional manual feature engineering techniques due to the end-to-end learning pipeline~\cite{fey2023relationaldeeplearninggraph, robinson2024relbench}.
% \czk{this may not be accurate if checking Figure 3 of the original relbench paper, the main improvement is less time in feature engineering}

% While an effective pre-training paradigm can potentially accelerate the fine-tuning process 
% % \czk{this is vague since training can refer to both pre-training and fine-tuning} 
% due to better initialization, or even improve downstream performance due to pre-training on larger data, 
Designing pre-training objectives for RDL is particularly challenging.
% \czk{in terms of efficiency, maybe need to be careful since there's no experiment result on this part. I think the pre-training is more time-consuming than training using the task table} 
% \czk{I think a better way to motivate here is to discuss the challenge of pre-training for RDB instead of why MAE/CL cannot work. We may summarize 2-3 challenges, and in the main contributions part our method address these challenges correspondingly.}
% \jt{how this related to the need of a SSL???}
A single RDB typically supports multiple tasks, each definable by distinct SQL queries. Meanwhile, 
% \jt{for which methods there is no theoretical guarantee? or we should say "Due to potential differences in label semantics, it is hard to guarantee that ..... " }
% there is no theoretical guarantee that the latent representation obtained during the pre-training stage is relevant to the downstream task due to potential differences in label semantics.
due to potential differences in label semantics, it is hard to guarantee that the latent representation obtained during the pre-training stage is relevant to downstream tasks, which leads to unreliable pre-trained models.
Moreover, downstream tasks depend on temporal dynamics (e.g. next-window properties), which cannot be simply modeled by just pre-training on unlabeled input data alone.
% , or often demand specific temporal sampling with mild assumptions on semantic similarity between two consecutive events \cite{liu2024capturingtemporal}. 
A pre-training framework is desired to model the dynamics in the latent representation intrinsically without having to rely on downstream fine-tuning for predictive modeling. In addition, what makes RDL distinct is the dependency of tasks on relational schema graphs, which leads to growing complexity on complicated RDBs.
% While existing SSL methods, such as masked autoencoders \cite{he2022masked} and contrastive learning \cite{chen2020simple}, \rewr{often serve as pre-training objectives} and have achieved notable success in fields like computer vision and natural language processing, their transferability to RDL remains uncertain. 
Recently, \citet{liu2023flaky} showed that adapting traditional SSL methods \cite{he2022masked, chen2020simple} to RDBs can inadvertently capture spurious yet seemingly informative patterns---so-called “interesting-looking noises”---that undermine downstream performance. This contradicts the core objective of traditional SSL, which aims to retain task-relevant information while discarding irrelevant noise \cite{huang2025learning, liang2023factorized, achille2018emergence, federici2020, alemi2017deep}, based on the assumption that different augmented views still share the same labels. Accordingly, our goal is to design a pre-training objective for RDL that avoids these pitfalls.
% \jt{why this happens for RDL???}
% RDL presents an unprecedented challenge: relevant information for one downstream task may be irrelevant for others.
% \jt{similarly why RDL needs task relevance or temporal dependencies differently from traditinal data which can work with reconstruction or contrastive separation }

\paragraph{Main Contributions} Our main contributions are summarized as follows:
\begin{enumerate}[%
    topsep=0pt,
    itemsep=0pt,
    leftmargin=15pt,
    label=\arabic*.  % optional: change label style
  ]
  \item \textbf{Task Vector Estimation (TVE)}: a pre-training framework for RDL that leverages schema graphs, next-window dynamics, and task heterogeneity to produce temporal, task-aware representations.
  \item \textbf{Theoretical insights}: we prove that pre-training on full relational databases with our objectives preserves more downstream-relevant information than standard self-supervised methods.
  \item \textbf{Empirical validation}: across diverse downstream tasks---especially in low-data regimes---TVE outperforms traditional SSL methods; ablation studies further uncover ``flaky'' task-dependent behaviors, underscoring the need to model both temporal dynamics and task heterogeneity.
\end{enumerate}

%% file: sections/2_Preliminaries.tex
\section{Preliminaries} \label{preliminaries}

% \jt{formally intro relational database}
% \jt{graph view of relational database}

% \czk{this part may be separated into the following four parts.}

% \czk{\textbf{Relational database~\cite{fey2023relationaldeeplearninggraph}.}}

% \czk{\textbf{Relational database as a temporal heterogeneous graph.} This includes a schema graph and an entity graph...}

% \czk{Can add some discussion between RDB and heterogeneous graph, so heterogeneous graph pre-training can not directly be applied to RDB.}

% \czk{\textbf{Predictive task over RDB.}  }

\textbf{Relational database \cite{fey2023relationaldeeplearninggraph}~} RDB \((\mathcal{T}, \mathcal{L})\) is formally defined as a set of tables (or entity types) \(\mathcal{T} = \{T_1, \dots, T_n\}\) that are connected by a set of PK-FK relations (or edge types) \(\mathcal{L} \subseteq \mathcal{T} \times \mathcal{T}\) \cite{fey2023relationaldeeplearninggraph}. Each table \(T_i\) can contain different data types for each column: numerical, categorical, text embeddings, etc. Additionally, each table contains a key column, which is either for primary keys or foreign keys, serving as links between tables. Tables are either \textit{fact} or \textit{dimension} tables \cite{fey2023relationaldeeplearninggraph}. Fact tables contain interaction between entities (e.g. \(T_{\text{review}}\)) with corresponding timestamps, while dimension tables contain static, immutable properties of entities (e.g. \(T_{\text{customer}}\) or \(T_{\text{product}}\)). Additionally, we refer root table to the table we would like to perform predictive modeling.
% \czk{this seems too compact, may need to add some details on what's the content of each table}

% \czk{Here I think we may have a paragraph called "graph perspective of RDB" and schema graph and entity graph are two parts.}
\textbf{Graph perspectives of RDBs.~} RDB can be understood from two different perspectives, namely schema graph and relational entity graph. \textbf{Schema graph} \((\mathcal{T}, \mathcal{R})\) is a blueprint of the RDB, in which the relation is bi-directional \(\mathcal{R} = \mathcal{L} \cup \mathcal{L}^{-1}\) \cite{fey2023relationaldeeplearninggraph}. \textbf{Relational entity graph} is the input graph for learning, and is a fine-grained view of the schema graph: we treat rows in a table as nodes, and they are connected to rows in other tables via the PK-FK links \cite{fey2023relationaldeeplearninggraph}. 
We use “nodes” and “rows” interchangeably in this work. We are interested in the relational entity graph snapshot up to a cutoff timestamp \(t\): 
% We consider node-level prediction tasks—either classification or regression—over temporal heterogeneous graphs constructed from relational databases.
{
\begin{equation}
    \mathcal{G}^{(-\infty, t]} = (\mathcal{V}^{(-\infty, t]}, \mathcal{E}^{(-\infty, t]}, \varphi_\mathcal{V}, \varphi_{\mathcal{E}}, f_{\mathcal{V}}, f_{\mathcal{E}}),
\end{equation}
}

where \(\mathcal{V}^{(-\infty, t]}\) and \(\mathcal{E}^{(-\infty, t]}\) are the sets of nodes and edges observed up to timestamp \(t\), \(\varphi_\mathcal{V}: \mathcal{V} \to \mathcal{T}\) maps each node to an entity type, \(\varphi_\mathcal{E}: \mathcal{E} \to \mathcal{R}\) maps each edge to a PK-FK link type, and \(f_{\mathcal{V}}: \mathcal{V} \to \mathbb{R}^{d}\) and \(f_{\mathcal{E}}: \mathcal{E} \to \mathbb{R}^{d}\) map nodes and edges to their feature vectors, respectively.

% The underlying \jt{do people know what is a schema graph? I think we need one definition either by us or from other papers if used} schema graph is denoted by \((\mathcal{T}, \mathcal{R})\) where \(\mathcal{T}\) is the set of entity types (i.e., tables), \(\mathcal{L}\) is the set of directed links (e.g., foreign key relations), and \(\mathcal{R} = \mathcal{L} \cup \mathcal{L}^{-1}\) is the edge set that includes both forward and backward links to account for bidirectional message passing. 

% \czk{Here we may add a short remark on difference between RDB to temporal graph and heterogenous graph, which renders SSL designed for them ineffective}

% \jt{I did not get the connection in these two sentences, please rewrite???}
Different tables have varying attribute sets, so node feature representations are inherently heterogeneous. Although edges could carry features, we use exclusively node features to simplify discourse.

\begin{figure}[t]
  \centering
  % give every subfigure a 4cm‐high box, align box bottom, and contents bottom:
  \begin{subfigure}[b][4cm][b]{0.2975\linewidth}
    \centering
    \includegraphics[width=\linewidth]{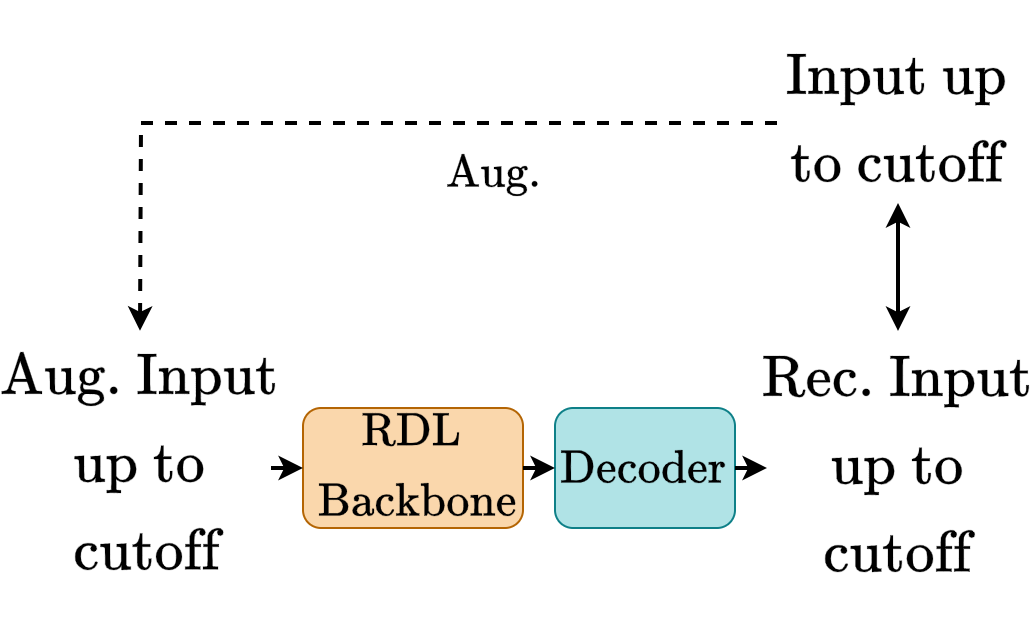}
    \caption{Masked Autoencoder}
    \label{subfig:mae}
  \end{subfigure}\hfill
  \begin{subfigure}[b][4cm][b]{0.33\linewidth}
    \centering
    \includegraphics[width=\linewidth]{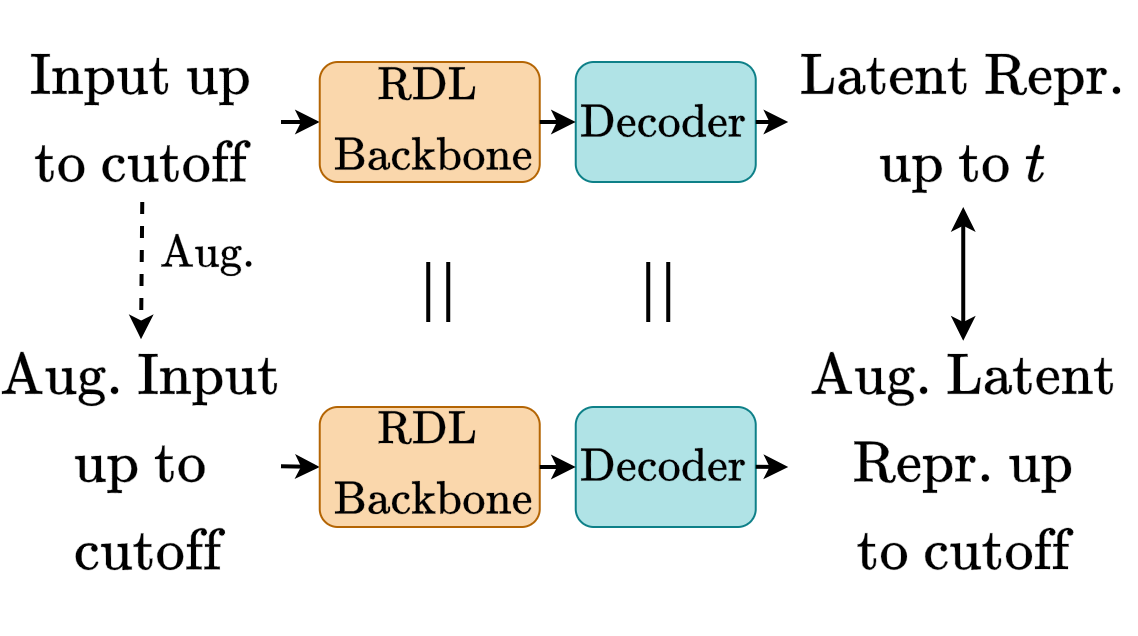}
    \caption{Contrastive Learning}
    \label{subfig:cl}
  \end{subfigure}\hfill
  \begin{subfigure}[b][4cm][b]{0.3375\linewidth}
    \centering
    \includegraphics[width=\linewidth]{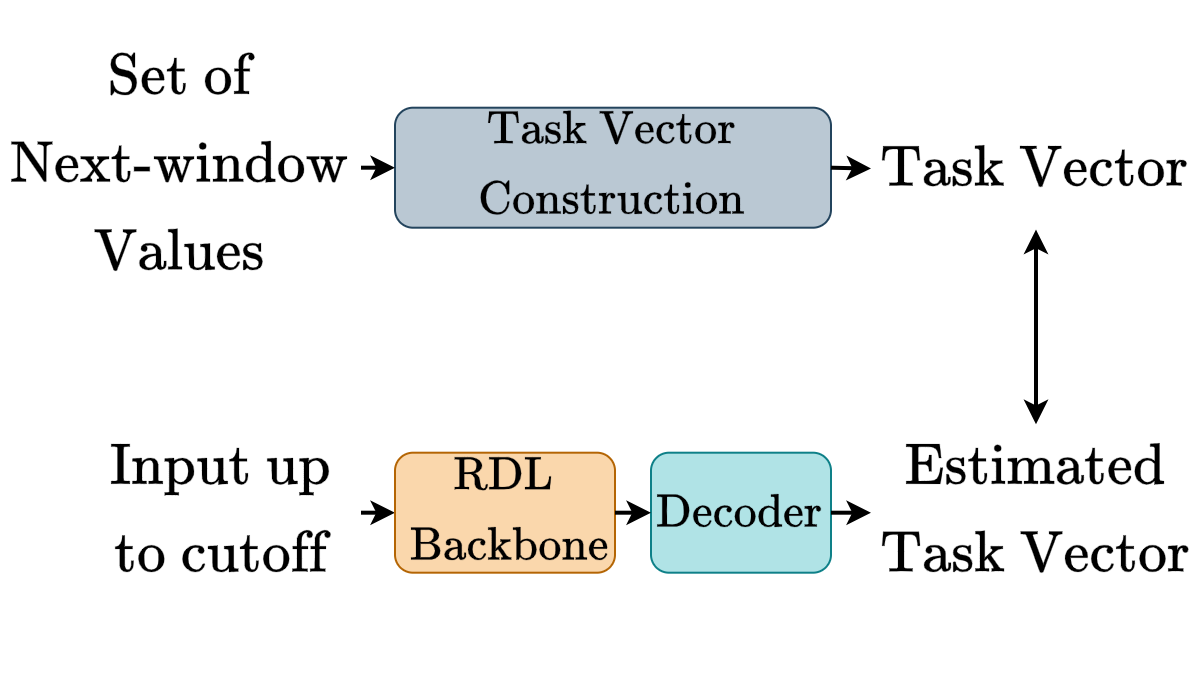}
    \caption{Task Vector Estimation}
    \label{subfig:tve}
  \end{subfigure}

  \caption{Overview of pre‐training frameworks. Input graph is omitted for brevity.}
  \label{fig:pre-training-frameworks}
\end{figure}

\textbf{Predictive tasks over RDBs.~} RDL spans a range of tasks, from table cell prediction \cite{wang20244dbinfer} to next-window property forecasting \cite{robinson2024relbench}. In this study, we concentrate on predictive modeling over relational data: given a training table \(T_\text{train}\), the objective is to forecast future node-level properties as formalized by RelBench \cite{robinson2024relbench}.
% \jt{we should write our paper in a more general way. RelBench could be one illustrative example?}
Each training node is associated with a labeled entity \(v = (\mathcal{K}_v, t_v, y_v^{(t_v, t_v + \Delta t]})\), where \(\mathcal{K}_v\) is a foreign key linking the entity to its originating table row, \(t_v\) is the cutoff timestamp, and \(y_v^{(t_v, t_v + \Delta t]}\) is the next-window label. 
% representing events occurring in the prediction window \((t_v, t_v + \Delta t]\). 
The learning objective is to estimate:
{
\begin{equation}
    \hat{y}_v^{(t_v, t_v + \Delta t]} = f_\theta(f_{\mathcal{V}}(v), \mathcal{G}^{(-\infty, t_v]}, t_v),
\end{equation}
}

where \(f_\theta\) is a differentiable function parameterized by \(\theta\) that uses the node's tabular features, the graph structure, and temporal context up to \(t_v\) to predict the future label. 
% Following prior work \cite{robinson2024relbench, fey2023relationaldeeplearninggraph}, \(f_\theta\) is instantiated as a heterogeneous GNN so that we can take into account of the heterogeneity of entities.

% \jt{we should say" next, we use xxxx as an example to illustate how to ..... }
Next, we would like to Fig. \ref{fig:tve-diagram}'s schema graph to demonstrate the process of forecasting customers’ review counts in the seven days following time $t$ as an example.
First, we extract \(\mathcal{G}^{(-\infty, t]}\) for learning. Second, we label each node by issuing an SQL query that joins the root table (e.g. \(T_\text{customer}\)) with with a fact table (e.g. \(T_\text{review}\)) and aggregates event counts over the interval \((t, t + 7 \ \text{days}]\). The node ID, the cutoff timestamp, and the corresponding label are recorded in the training table \(T_\text{train}\). This join-and-aggregate step is necessary because the root table is a dimension table, thus relying on neighboring fact tables to provide temporal contexts. 
% Another reason is to eliminate temporal leakage during training, where the sampled subgraphs are extracted according to the cutoff timestamp.
Third, follow \cite{fey2023relationaldeeplearninggraph, robinson2024relbench}, we train a model to predict these labels. \(T_\text{train}\) is used only for lookup and loss computation, not for message passing.

RDL is inherently distinct from both static heterogeneous and temporal graph problems. Static heterogeneous graph methods \cite{zhang2019heterognn, wang2019heterogat} omit temporal dynamics, while temporal graph approaches \cite{liu2024capturingtemporal, chen2023ssldynamicgraph} represent data as discrete-time snapshots or continuous-time event streams to capture graph evolution. By contrast, RDL treats timestamps as tabular features and uses them to impose causal constraints when extracting subgraphs. Lastly, RDL generates labels via SQL-driven schema‐graph joins, leveraging fact tables, aggregates, and multi‐hop joins absent from standard graph problems.
% Moreover, RDL defines links in a more flexible manner than simple pairwise events - \jt{do we test on any link prediction task? if not, we may use differetn examples} for instance, links may arise from fact tables, leading to different types of link prediction tasks for the same RDB. As a result, techniques developed for static or temporal graphs may not be directly applicable to the RDL setting.

%% file: sections/3_Pretraining_Framework.tex
% \begin{figure}[t]
%     \centering
%     \includegraphics[width=0.999\linewidth]{figures/framework.png}
%     \caption{Illustration of three pre-training frameworks that are studied in this paper: Masked Autoencoder, Contrastive Learning, and Task Vector Estimation. \(\mathcal{G}^{(-\infty, t]}\) is omitted from the figure for brevity.}
%     \label{fig:pre-training-frameworks}
% \end{figure}

\section{An Overview of the Proposed Pre-training Framework }\label{sec:overview}
% \czk{what's the relationship between section 3 \& 4. I feel they like two components of the whole pre-training framework. }
% \jt{problem statement}
% \jt{An overview}
% \qt{High-level pre-training}
% \czk{after reading the method part, I also find it a little bit hard to understand. Maybe these can help: 1. give a pseudocode for the whole process. 2. Maybe it's better to first introduce the method part itself, and then have an independent section telling about the theoretical motivations. }

% \jt{we need o rewrite this section. we should first present our frameowrk and introduce it. Then we can say we extend traditonal masked and contrastive learning to pretiraning in RDL. Last, discuss how our proposed one different from them and what are our advantages..}

% \jt{I have a concern about the figures. There are many notations in the figure not defined. Is it possible to replace these notations with descriptions? Can we use three subfigures to illustate these three methods. In this way, we can index them and it is easy to refer to during the main content.}

We would like introduce an overview of our pre-training framework, namely Task Vector Estimation (TVE), as shown in Fig. \ref{subfig:tve}. Besides the RDL backbone to extract features and the decoder to map the extracted latent representation to the final task vector, our framework contains an external module called Task Vector Construction to construct an objective vector extracted from the set of next-window values. By depending on temporal dynamics, schema graph, and simple SQL logics \cite{codd1970relational}, the vector represents a proxy representation for the set of next-window values---which plays an important roles in label generation process (see Section~\ref{subsec:label-generation}). Details are discussed in Section \ref{sec:predictive}, where we outline the motivation, implementation, and theoretical justifications.

% \jt{we should say: we also extent two of the most popular SSL frameworks to RDL and they are shown in figure 1 (a) and (b) directly. Them we can discuss how our frameork different from them and what are the advantages of our framework}

Most popular SSL frameworks, namely Masked Autoencoder (MAE) \cite{he2022masked} and Contrastive Learning (CL) \cite{chen2020simple}, are shown in Fig. \ref{subfig:mae} and Fig. \ref{subfig:cl}, respectively. MAE's objective is to reconstruct original feature values up to timestamp cutoff given augmentations, while CL attempts to minimize the distance between latent representations of two views of the same features (e.g. original features vs. corrupted features). Notably, the roles of decoders are different for MAE and CL. The decoder in MAE maps the extracted latent representation back to the original input space for reconstruction, whereas the decoder (often called the projection head) in CL maps the latent representation to a different, usually smaller space for contrastive learning. Both these methods depend on separation of data from inputs without modeling the next-window dynamics, which are critical for predictive modeling. Meanwhile, what sets TVE apart is the incorporation of an objective conditioned on next-window dynamics. TVE estimates task-relevant information during pre-training, and can be complementary to both MAE and CL (see Eq. \ref{eq:combined-loss} for the combined objective and Table \ref{tab:original-tasks} for experimental results). Our objective correlates input features and next-window dynamics, thus yielding a task-aware latent representation.

% \subsection{On the Connection between Latent Representation and Downstream Tasks} \label{subsec:many-tasks}

% \czk{I think this should be put together with 4.2. }

%% file: sections/4_Task_Vector_Estimation.tex
\section{Predictive Task-aware Self-supervised Learning}\label{sec:predictive}

In this section, we would like to shed light on the downstream label generation process, then propose our SSL method, namely \textbf{Task Vector Estimation (TVE)}, based on the label generation insights. Finally, we establish the benefits of this SSL objective over other methods from an information-theoretic perspective.

% \czk{My suggestion: First, provide an overview of the method here. To address the challenges mentioned in the introduction, we propose XXX. You can add some summary of motivation between the following sentences. First, it will generate a schema traversal graph to extract pre}

\subsection{Label Generation Process as a Set Function} \label{subsec:label-generation}

% \czk{better change a subtitle, this is too vague}

% \jt{in Section 2, after we intro the general notatins, can we use one example to illustrate the whole process.  Then the following discussions make more sense. Otherwise, it is super unclear}

\begin{figure}[t]
    \centering
    \includegraphics[width=0.85\linewidth,trim={0.4cm 0.55cm 0.4cm 0.3cm}, clip]{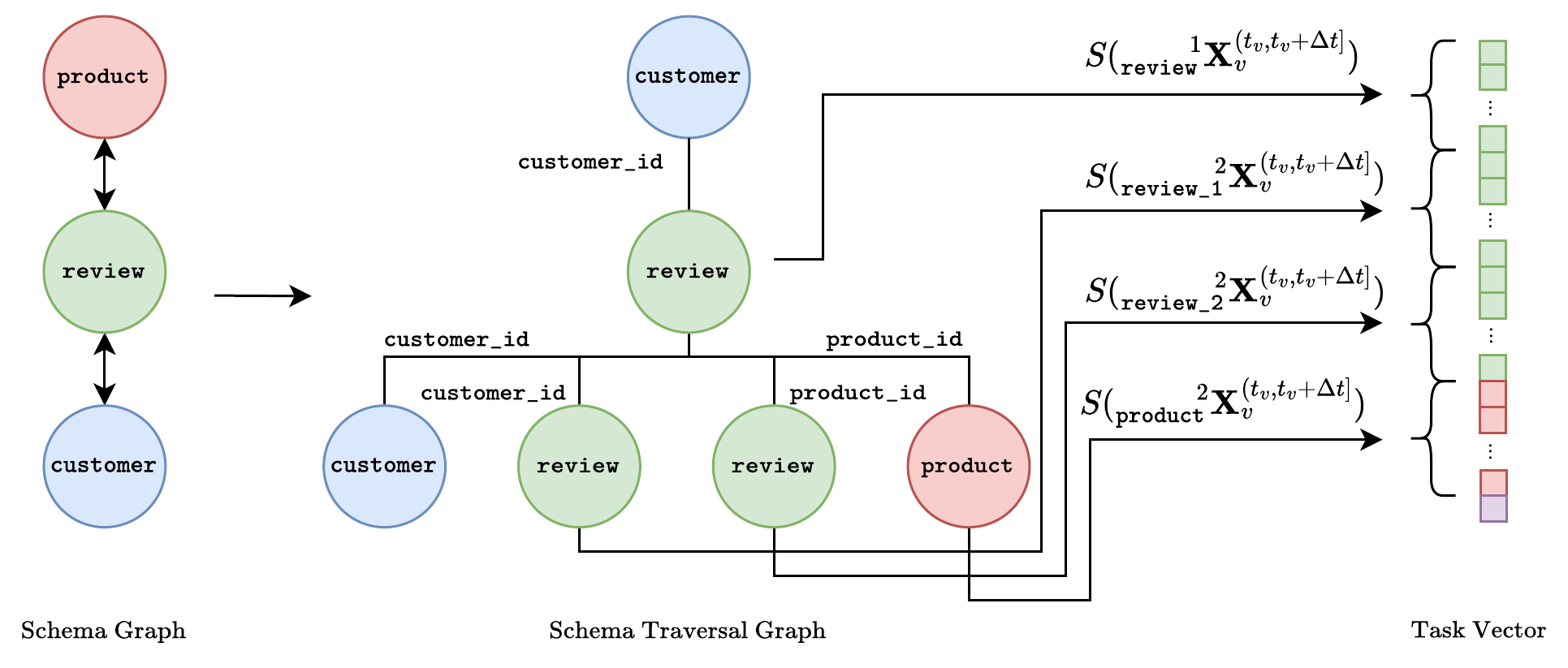}
    \caption{Construction of task vectors from relational schema. From schema graph, we derive a schema traversal graph, where edges represent valid join keys. For each reachable table \(R\) in \(k\)-hop from the source node \(v\), we apply a set function \(S\) to the set of next-window values \({}_R^k\mathbf{X}_v^{(t_v, t_v + \Delta t]}\) reachable via the traversal path, where only paths contributing to the prediction task are retained.
    % Only paths contributing to the prediction task are retained; for instance, 2-hop paths ending at table \texttt{customer} are excluded because \texttt{customer} is the root table. 
    The task vector concatenates outputs from these set functions. A null indicator (purple cell) is activated when all other entries are absent (i.e., zero-valued).}
    \label{fig:tve-diagram}
\end{figure}

Given the nature of labels defined in the training table introduced in Section~\ref{preliminaries}, we make the following observations regarding label generation process.  First, the training process is entity-centric---where we aim to perform predictive modeling for entities in the root table. Second, downstream tasks span across multiple tables in the RDB, and are conditioned on a set of nodes connected to the root table.
% \czk{is it a single table? do you mean the output is one single table? SQL is conducted over multiple tables}
Third, because the root table is often a dimension table, assigning a cutoff timestamp \(t_v\) requires 
% \jt{I did not get the following observation from Section 2}
joining it with fact tables. For example, tasks related to the \texttt{customer} table are timestamped via interactions with the \texttt{review} table, which is one hop away in the schema graph as shown in Fig. \ref{fig:tve-diagram}.

% \jt{I did not get the meaning of the following sentnece? this structure is what??} 

% \rewr{As labels are dependent on both the schema graphs and semantics of columns of tables in a path, we can have a broad space of downstream label.} One might, for example, define a task predicting how many reviews a customer submits in the prediction window \((t_v, t_v + \Delta t]\), which requires a 1-hop join involving \texttt{customer} and \texttt{review}, or compute the average price of products reviewed by the customer, which requires a 2-hop join involving \texttt{customer}, \texttt{review} and \texttt{product}. Formally, downstream labels can be represented as:

% \czk{The general writing may be polished using some tools. Here, some clauses like "in this case" appear multiple times consecutively.}
The set of next-window values depends not only on temporality, but also the schema graph structure and SQL logics. For example, predicting the number of reviews a customer will submit in the next window requires a one-hop join between the \texttt{customer} and \texttt{review} tables, with the \texttt{COUNT} aggregation. Likewise, computing the average price of products reviewed by a customer in that same window involves a two-hop join spanning \texttt{customer}, \texttt{review}, and \texttt{product}, coupled with the \texttt{AVERAGE} aggregation. For the sake of simplicity, assume we only consider the \(k\)-hop neighbors with the same entity-type, given an entity \(v\) and the timestamp \(t\), its label \(y_v^{(t)}\) is the output of a set function \(l\) defined over the features of its set of \(k\)-hop neighbors \(\mathbf{X}_v^{(t + 1)}\) in the next timestamp \(t+1\):
{
\begin{equation}\label{eq:set-function}
y_v^{(t)}  = l(\mathbf{X}_v^{(t + 1)})
  ,\quad
\mathbf{X}_v^{(t + 1)}  = \bigl\{f_{\mathcal V}(u)\mid u\in\mathcal{N}_k^{(t+1)}(v)\bigr\}.
\end{equation}
}

The formal definition of Eq.~\ref{eq:set-function}, where we additionally differentiate sets of neighbors conditioned based on their entity type \(R\), is included in Appendix~\ref{appendix:formal-def}. Importantly, the traversal graph permits self-joins, e.g., joining entries in \texttt{review} by both \texttt{customer\_id} and \texttt{product\_id}. When the labeling function \(l\) is an identity function and sets contain a single element, Eq.~\ref{eq:set-function} reduces to standard cell prediction, as often studied in Tabular Deep Learning~\cite{gorishniy2021revisitingdltabular, rubachev2022revisitingpretrainingobjectivestabular, ucar2021subtab, yoon2020vime, bahri2022scarf}.

\subsection{Task Vector Estimation}\label{subsec:tve}

% \czk{My suggestion is to not start from that SSL fails, but start from the challenge of RDB pre-training we summarize in the introduction}

% Conventional SSL paradigms \rewr{often} assume that \jt{not necessary with augmented views for SSL, let us be carefully in our claim} \rewr{corrupted} views of the input retain task-relevant features~\cite{huang2025learning, liang2023factorized, shen2024beyond, federici2020, tian2020whatmakes}. However, in RDL, downstream tasks may depend on disjoint or conflicting subsets of columns, \rewr{rendering such assumptions invalid if the pre-trained models are biased towards certain features only ~\cite{liu2023flaky}}. Furthermore, these paradigms typically ignore the next-window dynamics (see Fig. \ref{fig:pre-training-frameworks}), which are crucial for \rewr{predictive modeling}.

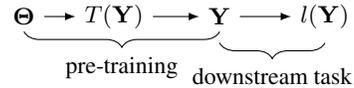
\begin{wrapfigure}{r}{0.5\textwidth}
  \centering
  \vspace{-1em}               % tighten up the top
  \begin{tikzpicture}[
    baseline=(current bounding box.center),
    every node/.style={font=\small},
    >=latex
  ]
    % nodes
    \node (θ)  at (0,0)   {$\mathbf\Theta$};
    \node (T)  at (1.2,0) {$T(\mathbf Y)$};
    \node (Y)  at (2.6,0) {$\mathbf Y$};
    \node (lY) at (4.0,0) {$l(\mathbf Y)$};

    % arrows
    \draw[->] (θ) -- (T);
    \draw[->] (T) -- (Y);
    \draw[->] (Y) -- (lY);

    % braces
    \draw[decorate,decoration={brace,mirror,amplitude=6pt}]
          (0,-0.2) -- (2.6,-0.2) node[midway,below=6pt]{pre-training};
    \draw[decorate,decoration={brace,mirror,amplitude=6pt}]
          (2.6,-0.35) -- (4.0,-0.35) node[midway,below=6pt]{downstream task};
  \end{tikzpicture}
  \vspace{0.5em}               % tighten up the bottom
  \caption{Markov Chain of Labels.}
  \label{fig:markov}
\end{wrapfigure}

As demonstrated in Section \ref{subsec:label-generation}, there are three key factors governing the label generation process: schema graphs, temporal dependencies, and SQL logics. While these three factors can complicate possible tasks, all tasks depend on  \({}^k_R\mathbf{X}_v^{(t_v, t_v + \Delta t]}\), which are represented by all possible rows, possibly with different entity types \(R\), that are \(k\)-hop from the root tables and associated with the next-window period \((t_v, t_v + \Delta t]\).

TVE's goal is to model task heterogeneity via estimating statistics representing \({}^k_R\mathbf{X}_v^{(t_v, t_v + \Delta t]}\). Generally, let \textbf{Y} represent a next-window set of values (e.g. \({}^k_R\mathbf{X}_v^{(t_v, t_v + \Delta t]}\)). Our goal is to capture the distributional properties of \(\vb{Y}\), parameterized by \(\mathbf{\Theta}\), as illustrated in Fig.~\ref{fig:markov}. Here, \(T(\mathbf{Y})\) denotes a sufficient statistic of \(\mathbf{Y}\) for \(\mathbf{\Theta}\). If the latent representation fully captures \(T(\mathbf{Y})\) during pre-training, then the latent representation has all information regarding \(\mathbf{Y}\) for \(\mathbf{\Theta}\) \cite{cover2006elements}. Finally, the downstream labels \(l(\mathbf{Y})\) depend on \(\mathbf{Y}\) as illustrated in Fig.~\ref{fig:markov} and formally defined in Section~\ref{subsec:label-generation}. Thus, pre-training model to estimate \(T(\mathbf{Y})\) yields a task-aware knowledge for downstream tasks.
% \[
% \begin{tikzpicture}[baseline=(current bounding box.center)]
%     \node (theta) at (0,0) {\(\mathbf{\Theta}\)};
%     \node (t) at (1.5,0) {\(T(\mathbf{Y})\)};
%     \node (y) at (3.5,0) {\(\mathbf{Y}\)};
%     \node (ly) at (5,0) {\(l(\mathbf{Y}),\)};
    
%     % Arrows
%     \draw[->] (theta) -- (t);
%     \draw[->] (t) -- (y);
%     \draw[->] (y) -- (ly);

%     % Braces at different heights
%     \draw[decorate,decoration={brace,mirror,amplitude=7pt}] 
%         (0,-0.2) -- (3.5,-0.2) node[midway,below=10pt] {pre-training};
    
%     \draw[decorate,decoration={brace,mirror,amplitude=7pt}] 
%         (3.5,-0.6) -- (5,-0.6) node[midway,below=10pt] {downstream task};
% \end{tikzpicture}
% \]

% Moreover, \jt{do we have any support to this claim?} modeling \(\mathbf{Y}\) during pre-training potentially reduces the latent code length required to support multiple tasks as conditional entropy of all tasks given a prior is at most as equal to the joint entropy without the prior.

TVE consists of two steps:
\begin{enumerate}[%
    topsep=0pt,
    itemsep=0pt,
    leftmargin=15pt,
    label=\arabic*.  % optional: change label style
  ]
    \item Generate a pre-training table by performing a \texttt{CROSS JOIN} between entity rows and candidate timestamps, then filtering out timestamps prior to the entity's first fact entry (e.g., first review).
    \item Attach pretext labels by computing normalized set-level statistics \(S\) (mean, count, etc.) per column across all reachable \({}^k_R\mathbf{X}_v^{(t_v, t_v + \Delta t]}\), using SQL aggregation.
\end{enumerate}

Regarding the first step, the intuition is that non-events entries are also important for downstream tasks such as churn prediction. The second step is illustrated in Fig. \ref{fig:tve-diagram}.  Specifically, given a relational schema graph, we construct a schema traversal graph based on joinability between two tables via PK or FK, in which edges denote which key is used for joining. Paths in this graph possess semantic meanings. Given a customer, we can find the set of next-window reviews made by the customer \(S( {}^{\quad \quad 1}_{\texttt{review}}\mathbf{X}_v^{(t_v, t_v + \Delta t]})\), the set of next-window products \(S( {}^{\quad \quad \ \ 2}_{\texttt{product}}\mathbf{X}_v^{(t_v, t_v + \Delta t]})\), the set of reviews for all products made by the customer \(S( {}^{\quad \quad \ \ \ \ 2}_{\texttt{review\_1}}\mathbf{X}_v^{(t_v, t_v + \Delta t]})\), and the set of reviews made by other customers for next-window products that the customer reviewed \(S( {}^{\quad \quad \ \ \ \ 2}_{\texttt{review\_2}}\mathbf{X}_v^{(t_v, t_v + \Delta t]})\). Importantly, the extracted set must be causal, which means that the joined entries in the next hop must have the timestamp less than or equal to that of the entries in the previous hop.
% Therefore, we can self-join an entry with another entry 
% % \jt{I think we should describe some details of figure 2 when we mention it otherwise, readers will be confused by what is review, customer id.... } 
% in table \texttt{review} by either \texttt{customer\_id} or \texttt{product\_id}.
Then, we aggregate values in sets by simple SQL aggregators to get predictive pretext tasks. 
% \jt{I do not understand the following sentence. Please justify why?} 
Note that we adopt the simple statistic \(S(\vb{Y})\) for computational efficiency, while approximating \(T(\mathbf{Y})\) is challenging. For further discussion regarding the sufficiency of simple statistics and future works on designing better approximations of sufficient statistics, please kindly refer to Appendix~\ref{app:further}.
% \jt{why for the following claim?}
Our proposed task vector construction operates similarly to Breadth First Search, and tasks extracted from a schema path is concatenated to the final task vector, leading to a variable-length task vector depending on the traversal paths. The task vector summarizes statistics for every column of tables reachable within \(k\)-hops of the root table in the schema traversal graph, estimating this vector ensures that no column is prematurely excluded during pre-training. It is an important design choice because downstream tasks may depend on any columns of any table. Optionally, the vector can be compressed via dimensionality reduction techniques such as PCA \cite{Pearson1901pca} or MCA \cite{abdi2007multiple}. The pre-training loss is the Scaled Cosine Error~\cite{hou2022graphmae}:

{
\begin{equation}\label{eq:tve-loss}
    \mathcal{L}_{\text{SCE}} = \frac{1}{|T_{\text{train}}|}\sum_{v \in T_{\text{train}}} \left( 1- \frac{\hat{y}_v^Ty_v}{||\hat{y}_v||\cdot ||y_v||} \right)^\alpha, \quad\alpha \geq 1, 
\end{equation}
}

where \(y_v\) is the ground-truth task vector, \(\hat{y}_v\) is the estimation of \(y_v\), and \(\alpha\) is a scaling factor. Additionally, to account for the potential imbalance due to null task vectors outnumbering others---we reweight each group’s contribution to the loss inversely proportional to its relative frequency.
% \jt{please explain any notations in the equation if we do not define them before??} 

Optionally, to mitigate the potential loss of input-specific information during pre-training, we extend our objective beyond the TVE loss (Eq.~\ref{eq:tve-loss}) by combining a traditional self-supervised loss:

{
\begin{equation} \label{eq:combined-loss}
    \mathcal{L} = \mathcal{L}_{\text{TVE}} + \beta\,\mathcal{L}_{\text{SSL}},
\end{equation}
}

where \(\beta\) is a balancing coefficient, and \(\mathcal{L}_{\text{SSL}}\) denotes a standard self-supervised loss---either a reconstruction objective or a contrastive loss. This formulation encourages the model to retain both predictive information (captured via task vectors) and general input features.

\subsection{Theoretical Evidence of Task Vector Estimation}\label{subsec:theoretical-evidence}

% \jt{we need to depict the high-level plans for the justification. Otherwise, we are confused about how we need the following??}

In this section, we would like to characterize relationship between 
% \jt{the latent representation produced by what??}
the latent representation of a pre-trained model and downstream tasks, and then we justify our design choice which can be proved to contain more task-relevant information than vanilla SSL methods.

% \jt{it is unclear about The broad range of tasks.  I suggest we term this challenge ealier "task hetergienirt??", here we can directly use it }
% The broad range of tasks introduces a unique challenge: 
Task heterogeneity introduces a unique challenge:
% \jt{we focus on SSL, we should rewrite the question about SSL instead of pretraining?? }
how to design a predictive task-aware SSL method for pre-training so that the model can generalize across a combinatorially large set of potential downstream tasks? Each task is conditioned on different column subsets of \({}^k_R\mathbf{X}_v^{(t_v, t_v + \Delta t]}\), and these tasks may or may not be correlated. This raises a natural question: how much task-relevant information must a latent representation encode to fully capture the entire set of tasks?
% \jt{to achieve what goal??? I feel the sentence is incomplete}

% \czk{it's weird to see the theory part starts with a remark statement. My suggestion is to start with a paragraph introducing the basic theoretical tool together with the notation system, then these claims can be made propositions.}

\begin{remark}\label{remark:latent-coding-scale}
    Suppose we have a set of tasks whose labels are denoted by \(\mathbf{y_1}, \mathbf{y_2}, \cdots, \mathbf{y_n}\). The lower bound of latent coding length \(L\) is:
    {
    \begin{equation}
        L \geq  H(\mathbf{y_1}, \mathbf{y_2}, \cdots, \mathbf{y_n})= \left[\sum_{i=1}^{n}H(\mathbf{y_i})\right] - TC(\mathbf{y_1}, \mathbf{y_2}, \cdots, \mathbf{y_n}),
    \end{equation}
    }
    
    where \(H(\mathbf{y}_i)\) is the entropy of \(\vb{y_i}\) and \(TC(\mathbf{y_1}, \mathbf{y_2}, \cdots, \mathbf{y_n})\) is total correlation of all tasks. Additionally, \(TC(\mathbf{y_1}, \mathbf{y_2}, \cdots, \mathbf{y_n}) \geq 0\) effectively becomes zero once all tasks are independent.
\end{remark}

The lower-bound of code length is covered in \cite{cover2006elements}, and the relationship between joint entropy and total correlation is studied in \cite{Watanabe1960tc}. Remark \ref{remark:latent-coding-scale} underscores an important property of any generalizable latent representation: the more uncorrelated tasks to support, the longer the required latent code. 
% Also note that there exist infinitely many tasks, which can be either dependent or independent from each other, as \({}^k_{R}\mathbf{X}_v^{(t_v, t_v + \Delta t]}\) depends on \(\Delta t\), which is discretized time. 
% Another noteworthy observation is that we can reduce the lower bound of the latent code length required to support multiple tasks if a prior \(\vb{t}\) is observed, as conditional entropy of all tasks given a prior is at most as equal to the joint entropy without the prior a.k.a \(H(\mathbf{y_1}, \mathbf{y_2}, \cdots, \mathbf{y_n}) \geq H(\mathbf{y_1}, \mathbf{y_2}, \cdots, \mathbf{y_n} \ | \ \mathbf{t})\).

Some tasks may degrade model performance. For example, if spending behavior is independent of gender, then training on gender-conditioned spending prediction tasks introduces noise. By the Data Processing Inequality, information can only decrease or remain the same through processing:

\begin{remark}
    Given the Markov chain \(\mathbf{y} \to \mathbf{x} \to \mathbf{x'}\) where \(\mathbf{y}\) is label, \(\mathbf{x}\) is original data, and \(\mathbf{x'}\) is the filtered data, then \(I(\mathbf{y}; \mathbf{x}) \geq I(\mathbf{y}; \mathbf{x'})\) by Data Processing Inequality.
\end{remark}

% Therefore, our goal is to design an SSL objective that leverages the full relational database, producing latent representations that act as a task-aware predictive code, since downstream labels are generated by applying set functions to future data's features as demonstrated in Section \ref{subsec:label-generation}.

Next, we provide an information-theoretic justification for TVE. Specifically, we would like to characterize how guiding models toward a predictive task-aware objective can benefit downstream tasks compared with vanilla contrastive separation or mask reconstruction from input data only.

Let \(\mathbf{x}\) denote input features, \(\mathbf{t}\) be a random variable representing side-channel information (which includes task vector proposed in Section \ref{subsec:tve}), and \(\mathbf{y}\) denote downstream labels. Let \(\mathbf{z}_1\) be a latent representation learned from both \(\mathbf{x}\) and \(\mathbf{t}\), and \(\mathbf{z}_2\) be learned from \(\mathbf{x}\) alone.

Prior literature \cite{federici2020} defines sufficiency of representation \(\vb{z_2}\) of \(\vb{x}\) for \(\vb{y}\), in which \(\vb{z_2}\) must be as predictive as \(\vb{x}\) to predict \(\vb{y}\). Similarly, we can define a condition for a sufficient representation of inputs and side channels for a downstream task.

% \begin{definition}[Sufficient representation given only inputs]\label{def:suff-z2}
%     A representation \(\vb{z_2}\) of \(\vb{x}\) is sufficient for \(\vb{y}\) if and only if \(I(\vb{x};\vb{y}|\vb{z_2}) = 0\).
% \end{definition}

\begin{definition}[Sufficient representation given inputs and side-channel information]\label{def:suff-z1}
    A representation \(\vb{z_1}\) is sufficient for predicting \(\vb{y}\) given input features \(\vb{x}\) with side-channel information \(\vb{t}\) if and only if \(I(\vb{xt};\vb{y} \mid \vb{z_1}) = 0\).
\end{definition}

% Sufficient representation of inputs and side-channel information for a task implies that the representation does not discard any task-relevant information contained in the inputs and side channels (see proof in Appendix \ref{proof:sufficient-condition}).

% % \jt{where are the proofs of the following proposiitons, theorem?..., we need to add index}

% \begin{proposition}\label{proposition:sufficient-condition}
%     Let \(\vb{x}, \vb{t}, \vb{y}\) be random variables with joint distribution \(p(\vb{x, t, y})\). Let \(\vb{z_1}\) be a latent representation of \(\vb{x}\) with additional side-channel information \(\vb{t}\). Then \(\vb{z_1}\) is sufficient of \(\vb{x}\) and \(\vb{t}\) for \(\vb{y}\) if and only if \(I(\vb{xt}; \vb{y}) = I(\vb{y}; \vb{z_1})\).
% \end{proposition}

% % \jt{where is the proof of proposition?? or simply justify how to get proposition}
% Consequently, the mutual information between the sufficient representation \(\vb{z_1}\) and the label \(\vb{y}\) is at least as large as that of the input \(\vb{x}\) and the label \(\vb{y}\) (see proof in Appendix \ref{proof:latent-contains-more}).

% \begin{corollary}\label{corollary:latent-contains-more}
%     Let \(\vb{x}, \vb{t}, \vb{y}\) be random variables with joint distribution \(p(\vb{x, t, y})\). Let \(\vb{z_1}\) be a sufficient latent representation of \(\vb{x}\) with additional side-channel information \(\vb{t}\) for \(\vb{y}\). Then, \(I(\vb{z_1}; \vb{y}) \geq I(\vb{x};\vb{y})\).
% \end{corollary}

% % \jt{where is the proof of corollary?? or simply justify how to get corollary}

We can prove that a representation of \(\vb{x}\) with additional knowledge of \(\vb{t}\) shares as least as much information with the downstream task than that of \(\vb{x}\) without \(\vb{t}\) does (see proof in Appendix \ref{proof:z1-better}).

\begin{theorem}\label{theorem:z1-better}
    Let \(\vb{x}, \vb{t}, \vb{y}\) be random variables with joint distribution \(p(\vb{x, t, y})\). Assume that \(\vb{z_1}\) is sufficient representation of \(\vb{x}\) with side-channel information \(\vb{t}\) for \(\vb{y}\), and \(\vb{z_2}\) is sufficient representation of \(\vb{x}\) for \(\vb{y}\). Then mutual information between \(\vb{z_1}\) and \(\vb{y}\) is at least as much as that between \(\vb{z_2}\) and \(\vb{y}\):
    \[
        I(\vb{z_1}; \vb{y}) \geq I(\vb{z_2}; \vb{y}).
    \]
\end{theorem}

% The equality holds when the side-channel information offers no additional insight about \(\vb{y}\) beyond what is already captured by \(\vb{x}\).  
% \jt{where is the proof of theorem?? or simply justify how to get theorem}
% The theorem suggests that incorporating side-channel information relevant to downstream tasks during pre-training can enhance the representational capacity of the model. This insight 
% % \jt{please rewrite the following, make it clear how the theorem can justify} 
% justifies the construction of the task vector, introduced in Section~\ref{subsec:tve}. Specifically, the task vector serves as a proxy representation for the set of next-window values associated with an entity type of interest. Since downstream labels are generated by heuristic set functions applied to these sets, the ability to reconstruct task vectors enables the model to approximate the underlying label-generating process more effectively.

The theorem justifies the construction of the task vector, introduced in Section~\ref{subsec:tve}. In other words, whenever side-channel signals actually carry extra task-relevant cues, using them in pre-training can equip the model with task-aware representations. The task vector is task-relevant, as it serves as the proxy representation of a set of values that are used during label generation process (as demonstrated in Eq. \ref{eq:set-function}), thus improving fine-tuning.
% Therefore, during fine-tuning, the model can better approximate labels more accurately because the model is aware of the underlying distribution of the set of values used to generate the labels. 
In contrast, standard SSL learns noise-invariant embeddings and only later fits a direct input--label mapping in the fine-tuning stage, without ever modeling the underlying set distribution in the pre-training stage.
% TVE closes this gap by training the encoder to forecast summary statistics of next-window value sets across all relevant columns, aligning the pretext task with the downstream labeling mechanism.

%% file: sections/5_Experiments.tex
\section{Experiments} \label{sec:experiments}

\begin{table}[t]
\caption{Performance across datasets and model variants in the low-data regime. Reported metrics in are
ROC-AUC for classification tasks, and Mean Absolute Error for regression tasks. Split, Validation, and Test are abbreviated as S, V, and T for readability. Bold is the best performance.}
\label{tab:low-data-tasks}
\setlength{\tabcolsep}{2.15pt}
\resizebox{\textwidth}{!}{%
\begin{tabular}{@{}lllccccccc@{}}
\toprule
\multicolumn{2}{l}{\multirow{2}{*}{Task}}                       & \multirow{2}{*}{S} & \multicolumn{7}{c}{Model}                                                                                                                       \\ \cmidrule(l){4-10} 
\multicolumn{2}{l}{}                                            &                    & Baseline      & MAE-0.25      & MAE-0.5       & CTR-0.25              & CTR-0.5               & TVE-1                  & TVE-2                  \\ \midrule
\multicolumn{10}{c}{Classification} \\ \midrule

\multirow{4}{*}{\makecell[tl]{rel-amz/\\item-churn}} & \multirow{2}{*}{\makecell[tl]{least\\50}}  & V                  & 76.04 ± 3.84  & 76.07 ± 4.38  & 76.06 ± 3.19  & 84.82 ± 2.12          & 84.32 ± 2.41          & 86.56 ± 1.31           & \textbf{87.11 ± 2.06}  \\
                                    &                           & T                  & 73.39 ± 6.63  & 71.33 ± 5.96  & 67.02 ± 6.89  & \textbf{83.97 ± 2.72} & 79.89 ± 4.41          & 81.97 ± 0.95           & 83.25 ± 2.40           \\ \cmidrule(l){2-10} 
                                    & \multirow{2}{*}{\makecell[tl]{least\\100}} & V                  & 76.64 ± 1.11  & 75.31 ± 2.05  & 74.03 ± 2.47  & 80.99 ± 1.27          & 80.09 ± 1.68          & 80.94 ± 1.35           & \textbf{82.79 ± 1.20}  \\
                                    &                           & T                  & 68.65 ± 1.91  & 68.48 ± 2.58  & 70.12 ± 1.58  & 74.78 ± 1.56          & 73.86 ± 1.68          & 75.29 ± 2.27           & \textbf{76.74 ± 1.46}  \\ \midrule
\multirow{4}{*}{\makecell[tl]{rel-amz/\\user-churn}} & \multirow{2}{*}{\makecell[tl]{top\\50}}    & V                  & 91.35 ± 1.27  & 91.53 ± 1.47  & 91.66 ± 1.13  & 91.01 ± 1.82          & 91.09 ± 1.10          & \textbf{93.31 ± 1.19}  & 91.82 ± 1.25           \\
                                    &                           & T                  & 89.07 ± 2.14  & 87.86 ± 2.00  & 85.71 ± 3.37  & 89.66 ± 2.26          & 89.71 ± 1.90          & \textbf{93.21 ± 1.58}  & 91.53 ± 1.84           \\ \cmidrule(l){2-10} 
                                    & \multirow{2}{*}{\makecell[tl]{top\\100}}   & V                  & 88.99 ± 0.61  & 90.00 ± 0.60  & 89.24 ± 0.68  & 91.32 ± 0.58          & 90.52 ± 0.86          & \textbf{91.35 ± 0.48}  & 90.92 ± 0.81           \\
                                    &                           & T                  & 86.23 ± 0.64  & 86.69 ± 1.00  & 85.59 ± 1.66  & 84.86 ± 1.39          & 86.32 ± 1.41          & 87.74 ± 1.04           & \textbf{88.13 ± 1.14}  \\ \midrule
\multirow{4}{*}{\makecell[tl]{rel-hm/\\user-churn}}  & \multirow{2}{*}{\makecell[tl]{top\\50}}    & V                  & 74.41 ± 0.85  & 76.30 ± 1.48  & 76.47 ± 1.20  & 84.07 ± 2.07          & \textbf{84.14 ± 1.74} & 76.47 ± 2.43           & 74.97 ± 1.42           \\
                                    &                           & T                  & 60.23 ± 1.84  & 56.94 ± 3.59  & 58.11 ± 2.38  & 56.82 ± 3.05          & 57.48 ± 3.55          & \textbf{65.01 ± 3.82}  & 64.12 ± 3.13           \\ \cmidrule(l){2-10} 
                                    & \multirow{2}{*}{\makecell[tl]{top\\100}}   & V                  & 72.63 ± 1.30  & 74.61 ± 1.48  & 76.22 ± 1.24  & 79.74 ± 0.67          & \textbf{79.85 ± 1.41} & 75.42 ± 1.53           & 75.60 ± 1.40           \\
                                    &                           & T                  & 63.43 ± 1.14  & 59.26 ± 1.77  & 60.25 ± 2.20  & 61.58 ± 1.15          & 60.69 ± 1.50          & 62.85 ± 1.87           & \textbf{64.06 ± 2.29}  \\ \midrule
\multicolumn{10}{c}{Regression} \\ \midrule
% \multirow{4}{*}{\makecell[tl]{rel-amz/\\item-ltv}}   & \multirow{2}{*}{\makecell[tl]{least\\50}}  & V                  & 0.085 ± 0.000 & 0.085 ± 0.001 & 0.085 ± 0.000 & 0.085 ± 0.001         & 0.084 ± 0.001         & \textbf{0.083 ± 0.001} & \textbf{0.083 ± 0.001} \\
%                                     &                           & T                  & 0.082 ± 0.000 & 0.082 ± 0.001 & 0.082 ± 0.000 & 0.081 ± 0.002         & 0.081 ± 0.001         & \textbf{0.078 ± 0.004} & 0.080 ± 0.003          \\ \cmidrule(l){2-10} 
\multirow{2}{*}{\makecell[tl]{rel-amz/\\item-ltv}} & \multirow{2}{*}{\makecell[tl]{least\\100}} & V                  & 0.175 ± 0.000 & 0.174 ± 0.001 & 0.172 ± 0.003 & 0.175 ± 0.001         & 0.174 ± 0.002         & 0.166 ± 0.005          & \textbf{0.165 ± 0.006} \\
                                    &                           & T                  & 0.186 ± 0.000 & 0.184 ± 0.003 & 0.185 ± 0.002 & 0.187 ± 0.000         & 0.188 ± 0.003         & \textbf{0.181 ± 0.005} & 0.188 ± 0.006          \\ \midrule
\multirow{2}{*}{\makecell[tl]{rel-amz/\\user-ltv}}   & \multirow{2}{*}{\makecell[tl]{bad-\\rev.}} & V                  & 8.172 ± 0.000 & 8.173 ± 0.000 & 8.172 ± 0.000 & 8.172 ± 0.000         & 8.172 ± 0.000         & 8.117 ± 0.008          & \textbf{8.115 ± 0.010} \\
                                    &                           & T                  & 9.520 ± 0.000 & 9.520 ± 0.000 & 9.520 ± 0.000 & 9.520 ± 0.000         & 9.520 ± 0.000         & \textbf{9.372 ± 0.026} & 9.373 ± 0.010          \\ \midrule
\multirow{2}{*}{\makecell[tl]{rel-hm/\\item-sales}}  & \multirow{2}{*}{\makecell[tl]{top\\200}}   & V                  & 3.203 ± 0.043 & 3.139 ± 0.083 & 3.111 ± 0.062 & 3.170 ± 0.090         & 3.113 ± 0.084         & 2.974 ± 0.098          & \textbf{2.945 ± 0.028} \\
                                    &                           & T                  & 3.240 ± 0.049 & 3.271 ± 0.120 & 3.249 ± 0.039 & 3.277 ± 0.135         & 3.284 ± 0.117         & \textbf{3.176 ± 0.078} & 3.205 ± 0.081          \\ \bottomrule
\end{tabular}%
}
\end{table}

\begin{table}[t]
\caption{Performance across datasets and model variants in the sufficient-data regime \cite{robinson2024relbench}. Reported metrics are
ROC-AUC for classification tasks, and Mean Absolute Error for regression tasks. Split, Validation, and Test are abbreviated as S, V, and T for readability. Bold is the best performance.
% \czk{Why the baseline performance here is lower than one in relbench}
} 
\label{tab:original-tasks}
\setlength{\tabcolsep}{2.15pt}
\resizebox{\textwidth}{!}{%
\begin{tabular}{@{}llccccccc@{}}
\toprule
\multirow{2}{*}{Task}               & \multirow{2}{*}{S} & \multicolumn{7}{c}{Model}                                                                                                                                  \\ \cmidrule(l){3-9} 
                                    &                    & Baseline       & MAE-0.25                 & MAE+TVE-1       & MAE+TVE-2       & CTR-0.25              & CTR+TVE-1                & CTR+TVE-2               \\ \midrule
\multicolumn{9}{c}{Classification} \\ \midrule
\multirow{2}{*}{\makecell[tl]{rel-amz/\\item-churn}} & V                  & 81.54 ± 0.06   & \textbf{81.61 ± 0.14}    & 81.13 ± 0.04    & 81.19 ± 0.03    & 81.27 ± 0.08          & 81.29 ± 0.09& 81.28 ± 0.10\\
                                    & T                  & 81.85 ± 0.10   & \textbf{81.92 ± 0.16}    & 81.47 ± 0.04    & 81.49 ± 0.04    & 81.53 ± 0.09          & 81.57 ± 0.13& 81.56 ± 0.12\\ \midrule
\multirow{2}{*}{\makecell[tl]{rel-amz/\\user-churn}} & V                  & 69.95 ± 0.09   & 69.26 ± 0.22             & 69.92 ± 0.04& 69.93 ± 0.05& 69.85 ± 0.05          & \textbf{70.04 ± 0.04}& 70.03 ± 0.06\\
                                    & T                  & 69.82 ± 0.05   & 68.89 ± 0.29             & 69.78 ± 0.07& 69.80 ± 0.03& 69.64 ± 0.09          & \textbf{69.95 ± 0.02}& 69.85 ± 0.07\\ \midrule
\multirow{2}{*}{\makecell[tl]{rel-hm/\\user-churn}}  & V                  & 70.34 ± 0.09& 70.39 ± 0.06             & 70.29 ± 0.09& 70.35 ± 0.10& 70.49 ± 0.13          & 70.53 ± 0.04& \textbf{70.54 ± 0.16}\\
                                    & T                  & 69.77 ± 0.33& 69.81 ± 0.20             & 69.89 ± 0.09& 69.80 ± 0.11& 69.97 ± 0.09& \textbf{70.13 ± 0.22}& 69.72 ± 0.51\\ \midrule
\multicolumn{9}{c}{Regression} \\ \midrule
\multirow{2}{*}{\makecell[tl]{rel-amz/\\item-ltv}}   & V                  & 45.818 ± 0.301& 45.573 ± 0.091           & 45.590 ± 0.121& 45.573 ± 0.045& 45.630 ± 0.161        & 45.030 ± 0.100& \textbf{44.956 ± 0.078}\\
                                    & T                  & 50.603 ± 0.351& 50.471 ± 0.308           & 50.437 ± 0.395& 50.659 ± 0.417& 50.472 ± 0.432        & 50.012 ± 0.287& \textbf{49.753 ± 0.192}\\ \midrule
\multirow{2}{*}{\makecell[tl]{rel-amz/\\user-ltv}}   & V                  & 12.416 ± 0.016 & 15.989 ± 1.154           & 12.429 ± 0.016& 12.437 ± 0.031& 12.434 ± 0.014        & 12.397 ± 0.010& \textbf{12.371 ± 0.018}\\
                                    & T                  & 14.665 ± 0.039 & 17.397 ± 0.907           & 14.658 ± 0.028& 14.626 ± 0.019& 14.687 ± 0.057        & 14.622 ± 0.046& \textbf{14.577 ± 0.020}\\ \midrule
\multirow{2}{*}{\makecell[tl]{rel-hm/\\item-sales}}  & V                  & 0.0643 ± 2e-4& 0.0630 ± 1e-4& 0.0627 ± 1e-4& \textbf{0.0624 ± 3e-4}& 0.0635 ± 2e-4& 0.0628 ± 4e-4& 0.0627 ± 3e-4\\
                                    & T                  & 0.0556 ± 2e-4& 0.0547 ± 1e-4& 0.0544 ± 2e-4& \textbf{0.0543 ± 3e-4}& 0.0554 ± 3e-4& 0.0547 ± 3e-4& \textbf{0.0543 ± 3e-4}\\ \bottomrule
\end{tabular}%
}

\end{table}

In this section, we conduct comprehensive experiments to answer the following research questions:

\begin{itemize}[%
    topsep=0pt,
    itemsep=0pt,
    leftmargin=10pt,
  ]
    \item RQ1: How does our proposed task vector improve model performance on downstream tasks?
    \item RQ2: Can TVE provide better latent representations compared to traditional SSL methods for downstream tasks?
    \item RQ3: How sensitive is our model to different hyperparameter settings and last linear head's initialization?
\end{itemize}

\subsection{Experimental Settings}

% \czk{need to add some reasons for selecting these two baselines. Maybe say there's no method specifically designed for RDB, while traditional method can be categorized in generative and discriminative, represented by MAE and CL.}
We compare TVE with other SSL methods that do not take into account next-window dynamics---Graph Masked Autoencoder and Graph Contrastive Learning. 
% Because no SSL pre-training techniques have been developed specifically for RDL - and existing SSL methods typically fall into generative or discriminative paradigms - we selected these two works to evaluate if noise-invariant objectives on feature matrices can serve as pre-text tasks.
Specifically, these architectures resemble GraphMAE \cite{hou2022graphmae} and GraphCL \cite{you2020graph}, with minor modifications for RDL pre-training. All evaluated models used the same backbone proposed by \cite{robinson2024relbench}. We compare several models: a non-pretrained Baseline; Graph Masked Autoencoder with mask rates of 0.25 and 0.5 (MAE-0.25, MAE-0.5); Graph Contrastive Learning with the same mask rates (CTR-0.25, CTR-0.5); and our proposed method, TVE, using 1-hop and 2-hop schema traversals (TVE-1, TVE-2). All pre-training is conducted using the same training table constructed via \texttt{CROSS PRODUCT} as described in Section~\ref{subsec:tve}. While there are many pre-training techniques designed for vertical domains, their translations to RDL are non-trivial due to the fundamental differences outlined in Section \ref{preliminaries}; therefore, we only consider pre-training techniques based on tabular feature perturbations (see Appendix~\ref{appendix:experiments} for details).
% \jt{here we should mention why we do not compare with pretrianing methods on hetergrnrtious and temporal graphs? like As metioned in Section 2, how RDL are dfiffernt from ....., so we do not choose pretrian..... }

We evaluate our proposed method on a variety of node-level tasks defined over two datasets: \texttt{rel-amazon} and \texttt{rel-hm}. Additionally, we introduce a suite of new tasks designed to reflect realistic scenarios. For instance, a business might wish to build predictive models for a specific cohort (e.g. low-spending customers) where label data is scarce. These scenarios often lack rich temporal context, so dynamics presented by labels alone may not support high-quality forecasting. Yet these tasks are crucial, since companies increasingly aim to personalize offerings for these groups to boost retention. To generate these tasks, we apply attribute‐based filters on the prior period: for instance, an SQL query retrieves the top-\(k\) spenders over the last three months, and other customers are excluded from the training data for the current period. 

% \jt{we still do not mention how to construct the datasets with limited labels??}
% \czk{need to strengthen these two are all different tasks with different training and test datasets, and discuss some real-world meanings of such tasks} 

To address this, we explicitly evaluate model performance under two regimes: (i) data limited tasks and (ii) data sufficient tasks (the original RelBench tasks). More details are outlined in Appendix \ref{appendix:experiments}.

% \czk{I think we need to put the table 2 first, if we claim improvement in table 1 while in table 2 the performance is not that good. Then the problem is that such pre-training hurt performance on popular user/items. }

\begin{figure}[t]
    \centering
    \includegraphics[width=\linewidth, trim={0cm 0.25cm 0cm 0cm},clip]{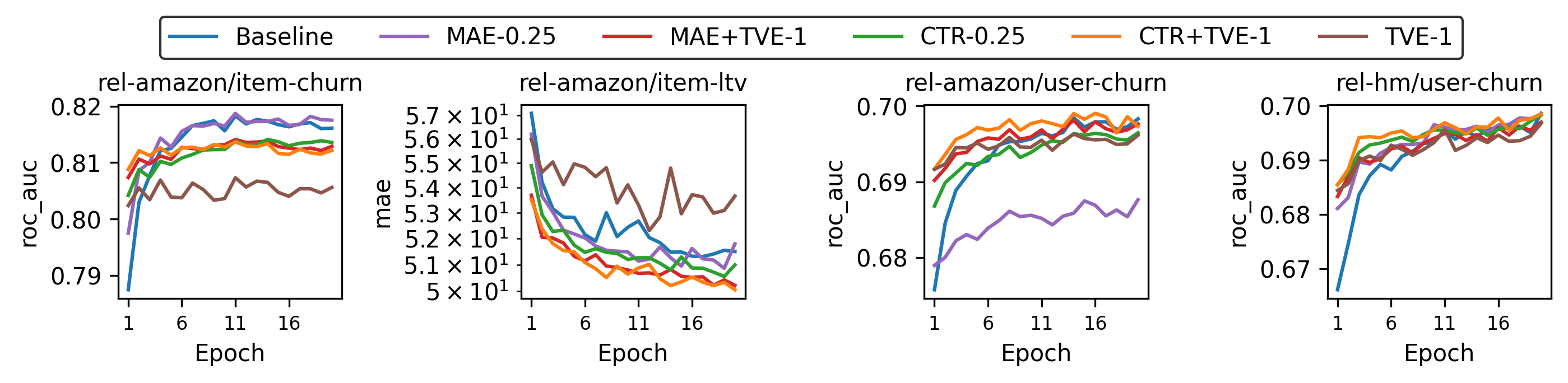}
    \caption{Test performance curves averaged over runs on data sufficient tasks.}
    \label{fig:sufficient-curves}
\end{figure}

\begin{figure}[t]
    \centering
    \includegraphics[width=\linewidth, trim={0cm 0.25cm 0cm 0cm},clip]{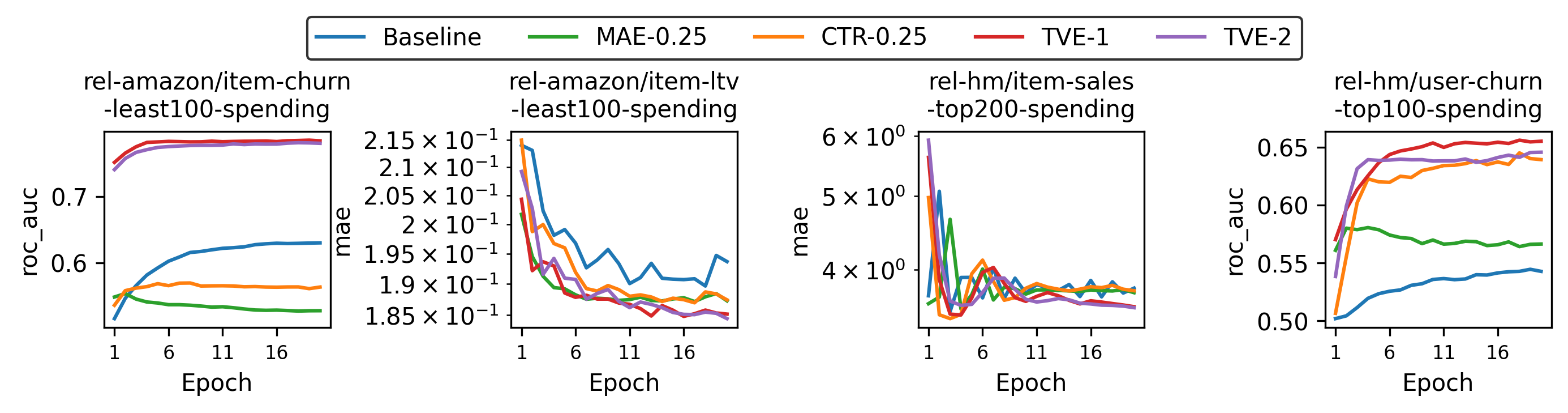}
    \caption{Linear probing results averaged over runs on test set.}
    \label{fig:linear-probing}
\end{figure}

\subsection{Data Limited Tasks} \label{subsec:data-limited}
Table~\ref{tab:low-data-tasks} presents results for tasks with limited labeled supervision. We find that TVE consistently outperforms other SSL baselines, achieving the best performance in 5 out of 6 classification tasks. The only exception is \texttt{rel-amazon/item-churn-least50-spending}, where CTR-0.25 achieves the highest test ROC-AUC. However, this task illustrates a form of \emph{flaky behavior}: while CTR-0.25 and CTR-0.5 exhibit nearly identical validation scores (84.82 ± 2.12 vs. 84.32 ± 2.41), their performance diverges significantly on test set (83.97 ± 2.72 vs. 79.89 ± 4.41). This suggests sensitivity to hyperparameter tuning (e.g., augmentation strength), which are later studied in Section~\ref{subsec:flaky}.

% More broadly, we observe that both MAE and CTR often achieve strong validation performance but underperform on the test set—sometimes even underperforming the Baseline. We hypothesize that these models are susceptible to distribution shift, as their pre-training objectives do not correlate inputs and next-window dynamics. In contrast, RelBench uses temporal splits for training, validation, and test sets, which implies that validation data is temporally closer to the training set, potentially causing recency bias. Our method explicitly captures next-window dynamics during pre-training, allowing it to generalize across time. Similar patterns hold in regression tasks: for instance, in \texttt{rel-amazon/user-ltv-bad-reviews}, MAE and CTR match Baseline performance, while TVE allows the model to achieve better performance due to predictive modeling during pre-training. These results answer RQ1, which demonstrates the effectiveness of TVE.

More broadly, MAE and CTR often excel on validation but falter on the test set---sometimes below the Baseline---likely due to distribution shift and recency bias from temporal splits. Our method explicitly captures next-window dynamics during pre-training, allowing it to generalize across time. Similar patterns hold in regression tasks: for instance, in \texttt{rel-amazon/user-ltv-bad-reviews}, MAE and CTR match Baseline performance, while TVE allows the model to achieve better performance. These results answer RQ1, which demonstrates the effectiveness of TVE.

% Additionally, Table \ref{tab:low-data-tasks} reveals that the optimal pre-training hop distance-i.e., the number of PK–FK traversals from the root table-differs across downstream tasks, underscoring that each task draws on information from a distinct subset of related tables. Even when the optimal tables lie beyond the reach of a fixed-hop neighborhood, our TVE models-which explicitly forecast next-window relational dynamics-still maintain strong performance. This robustness highlights the central role of temporally informed pre-training in capturing the predictive signals.

\subsection{Data Sufficient Tasks} \label{subsec:data-sufficient}

% \qt{Sufficient data is enough, but improvement is not significant. Explain why?
% Existing SSL complementary to TVE.
% Claim 2 things:
% 1. 
% need to claim all SSL isn't competent.
% our method is complementary.}

% In Table \ref{tab:original-tasks}, we benchmark each model on the original RelBench tasks, which involve substantially larger downstream datasets. In these data-rich scenarios, all SSL methods yield only marginal improvements over the baseline and vanilla TVE actually underperforms the other SSL variants on some tasks as illustrated in Fig.~\ref{fig:sufficient-curves}. We attribute this to the fact that, with abundant fine-tuning data, temporal patterns can be recovered during downstream training without explicit pre-training. This degradation in TVE’s performance is consistent with Theorem \ref{theorem:z1-better}, which asserts that our pre-training objective preserves additional downstream information only when the learned representation remains jointly sufficient for both the inputs and the side-channel task vectors; the observed drop, therefore, indicates that useful input information was lost during pre-training. As mentioned in Section \ref{sec:overview}, TVE is complementary to existing SSL methods. Therefore, we employ the combined-loss model from Eq. \ref{eq:combined-loss}, which penalizes the omission of original information. As Fig. \ref{fig:sufficient-curves} demonstrates, this hybrid objective produces a stronger initialization, and Table \ref{tab:original-tasks} confirms that it outperforms single-objective models on five of the six tasks.

In Table \ref{tab:original-tasks}, we benchmark each model on the original RelBench tasks, which involve substantially larger downstream datasets. In data‐rich settings, all SSL methods---including vanilla TVE---produce only marginal gains or even underperform, since plentiful fine-tuning data alone can recover temporal patterns. This matches Theorem \ref{theorem:z1-better}: TVE’s extra pre-training benefit arises only if the representation remains jointly sufficient for both inputs and task vectors, so any drop signals lost input information. As mentioned in Section \ref{sec:overview}, TVE is complementary to existing SSL methods. Therefore, we employ the combined-loss model from Eq. \ref{eq:combined-loss}, which penalizes the omission of original information. As Fig. \ref{fig:sufficient-curves} demonstrates, this hybrid objective produces a stronger initialization, and Table \ref{tab:original-tasks} confirms that it outperforms single-objective models on five of the six tasks.

% \subsection{Ablation Studies} \jt{ Linear Probing is not real ablation study, I suggest to make the following two subsub as two subsections }

\subsection{Linear Probing Performance} \label{subsec:linear-probing}
% \jt{I think RQ2 is not necessary, but this one is more imporant. Please replace it by "Can our method learn more task-specific information compared to ... }

To answer RQ2, we evaluate the quality of the learned representations independently of downstream fine-tuning by linear probing experiments \cite{zhang2016colorful}. Specifically, we freeze the pre-trained backbone and train only a linear head that maps latent representations to task outputs. All model shared the same backbone with the channel size of 128.
% Full experimental details are provided in Appendix~\ref{appendix:experiments}.

As shown in Fig.~\ref{fig:linear-probing}, representations learned via TVE outperform all other SSL methods, as well as the Baseline. 
% This result confirms that TVE captures task-relevant information more effectively during pre-training. 
Furthermore, CTR-0.25 achieves strong performance on three of four tasks, yet falls below the Baseline on \texttt{rel-amazon/item-churn-least100-spending}, illustrating the presence of ``interesting-looking noise'' \cite{liu2023flaky}. In contrast, TVE does not rely on augmentations, thus yielding stable and transferable embeddings. 
% Another takeaway is that while other SSL methods may have better initialization than Baseline, it does not guarantee that the end performance is better. 
% Meanwile, TVE can achieve both better initialization and better performance.

\begin{figure}[t]
    \centering
    \includegraphics[width=0.915\linewidth, trim={0cm 0.22cm 0cm 0.22cm},
  clip]{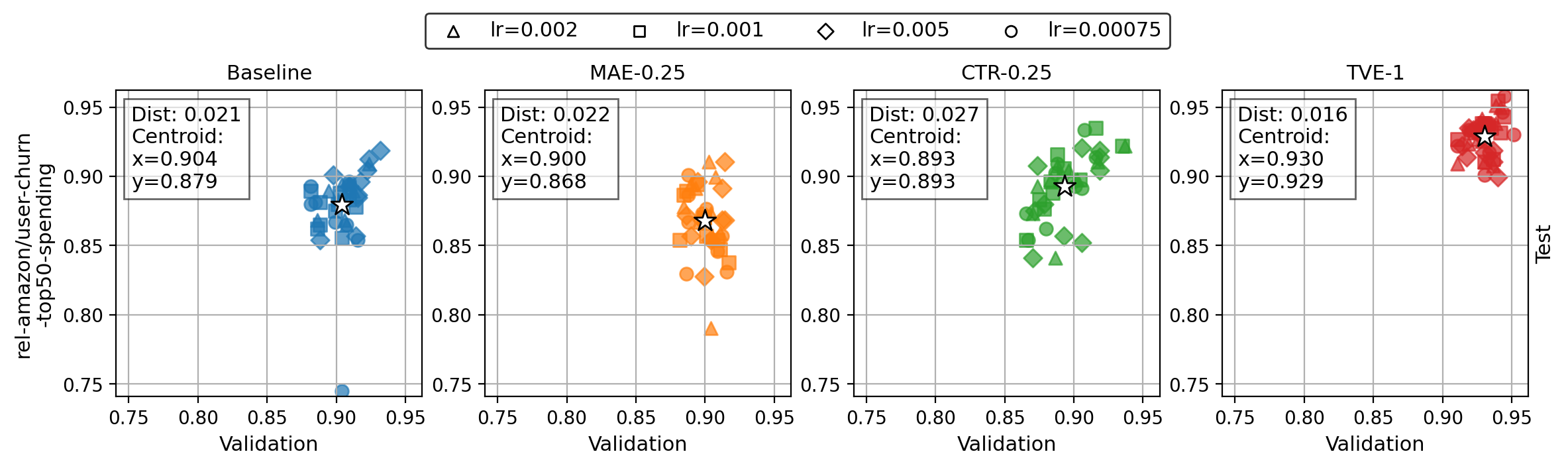}
    \caption{Validation vs. Test scatter plots for data limited node classification tasks across varying learning rates. Average distance to centroid (denoted \ding{73}) are used to measure performance variance.}
    \label{fig:flaky-scatter}
\end{figure}

\subsection{Performance Sensitivity Across Splits and Tasks}\label{subsec:flaky}

In order to answer RQ3, we further investigate the robustness of each method to hyperparameter variation by analyzing the stability of performance across learning rates and last linear head's random initialization. We vary the learning rate across 10 runs per task, yielding 40 data points per scatterplot in Fig.~\ref{fig:flaky-scatter}. 
% The average Euclidean distance to the centroid is used to quantify sensitivity. 
All models share the same backbone with the channel size of 128.

The results show that MAE-0.25 and CTR-0.25 are highly sensitive to hyperparameters and last linear head's random initialization: the spread of points is substantially larger compared to TVE. These results suggest that the task-aware, temporal pre-training of TVE leads to more stable and transferable representations in the low-data setting. Experiments for other tasks are provided in Appendix~\ref{appendix:experiments}.

%% file: sections/6_Discussion_and_Related_Works.tex
\section{Related Works, Discussion, and Conclusion} \label{sec:related}

% \jt{we certainly can reduce the related work if space is needed??}

\textbf{Relational Deep Learning.~}
RDL has emerged as a paradigm for learning from RDBs \cite{fey2023relationaldeeplearninggraph, cvitkovic2020supervisedlearningrelationaldatabases, robinson2024relbench, gan2024graphmachinelearning, wang20244dbinfer, bazhenov2024tabgraphsbenchmarkstrongbaselines}. 
% Any state-of-the-art tabular encoder—such as TabNet \cite{arik2021tabnet}, TabTransformer \cite{huang2020tabtransformer}, or FT-Transformer \cite{gorishniy2021revisitingdltabular}-and any modern graph encoder-e.g., GCN \cite{kipf2017semisupervised}, GraphSAGE \cite{hamilton2017inductive}, or GAT \cite{veličković2018deep}-can be employed to produce meaningful embeddings. In addition to tabular features and graph structures, \citet{fey2023relationaldeeplearninggraph} advocates temporal modeling as a first-class citizen when designing learning architectures for RDL. 
Although SSL techniques are advanced for individual domains---temporal graphs \cite{tian2021ssl, alomrani2024dygvec, chen2022pre, liu2024self, liu2024capturingtemporal, chen2023ssldynamicgraph}, static graphs \cite{veličković2018deep, hou2022graphmae, you2020graph}, and tabular data \cite{ucar2021subtab, yoon2020vime, bahri2022scarf}---SSL investigations on RDBs are still limited \cite{liu2023flaky}. 
% Pre-training objectives differ across vertical domains: static-graph methods assume local subgraph similarity or structural robustness \cite{veličković2018deep, you2020graph, hou2022graphmae}, temporal-graph methods impose consistency across successive events \cite{liu2024self, liu2024capturingtemporal}, and tabular methods typically rely on masking and reconstruction or contrastive losses \cite{bahri2022scarf, yoon2020vime, ucar2021subtab}. 
Drawing on \cite{nam2023stunt}’s \(k\)-means column-subset pretext tasks, TVE instead generates objectives via schema-graph traversals and temporal SQL aggregations. Unlike SQL-join feature synthesis \cite{kanter2015deep} or recent RDL advances on architectures \cite{zhang2023gfsgraphbasedfeaturesynthesis,yuan2025contextgnn} and efficiency \cite{lachi2025boostingrelationaldeeplearning}, our approach centers on task-space representations for SSL pre-training.

\textbf{Task Modeling.~}
Generalizing to new tasks remains a core challenge. Prior approaches include task taxonomies to reveal inter‐task structure \cite{Zamir_2018_CVPR}, task embeddings for selecting feature extractors \cite{Achille_2019_task2vec,cao2023autotransfer}, and modeling data–auxiliary task node interactions \cite{cao2023relational}. In contrast, we analyze label‐generation in relational deep learning and propose a pretext objective that explicitly preserves the next‐window set representations from which any downstream labels are derived.

\textbf{Representation Learning.~} Recent SSL–information‐theoretic work frames representations as noise‐invariant \cite{achille2018emergence} and studies how multi‐view SSL discards redundant features for downstream tasks \cite{federici2020}. Augmentations can erase view‐specific cues, underscoring the need to preserve task‐relevant signals \cite{liang2023factorized}. Another important SSL technique is to train models to predict future observations (e.g., frames or tokens) \cite{lotter2016deep,oord2018representation}. TVE follows these principles by forecasting next‐window dynamics to capture the core information for downstream labels.

\textbf{Limitations.~} Although our theory leverages sufficient statistics, task vectors---built via simple SQL aggregations---may not fully capture the original next‐window sets, thus leading to loss of information. Their dimensionality grows with the number of columns and joinable paths, which necessitates dimensionality reduction techniques \cite{Pearson1901pca, abdi2007multiple}. Finally, a learnable representation for next‐window sets may be more expressive; we leave this exploration to future work.

\textbf{Conclusion.~} We introduce Task Vector Estimation, a pre-training framework that integrates schema graphs, temporal dynamics, and task heterogeneity. Our theoretical results show that task-aware objectives retain more downstream-relevant information than conventional SSL methods, producing representations that are predictive for downstream tasks. Our findings underscore the importance of treating task heterogeneity and temporal context as first-class citizens in RDL pre-training strategies.
% Moreover, our use of SQL to craft complex pre-training signals highlights how database manipulations can uncover rich supervisory targets that benefit downstream performance.

%% file: sections/0_Appendix.tex
\appendix
\newpage
\section{Proofs} \label{appendix:proof}

In this section, we assume that all random variables \(\vb{x}, \vb{t}, \vb{y}, \vb{z}_1, \vb{z}_2\) are discrete. We use the following identities to prove the theorem reported in Section \ref{subsec:theoretical-evidence}. For any random variable \(\vb{x}\), \(\vb{y}\), and \(\vb{z}\), we have the following properties:

\textit{Non-negativity:}
\begin{equation}\label{eq:p1}
    I(\mathbf{x}; \mathbf{y}) \geq 0, \quad I(\mathbf{x}; \mathbf{y} \mid \mathbf{z}) \geq 0.
\end{equation}

\textit{Chain rule:}
\begin{equation}\label{eq:p2}
    I(\mathbf{x}\mathbf{y}; \mathbf{z}) = I(\mathbf{y}; \mathbf{z}) + I(\mathbf{x}; \mathbf{z} \mid \mathbf{y}).
\end{equation}

\textit{Chain rule (Multivariate Mutual Information):}
\begin{equation}\label{eq:p3}
    I(\mathbf{x}; \mathbf{y}; \mathbf{z}) = I(\mathbf{y}; \mathbf{z}) - I(\mathbf{y}; \mathbf{z} \mid \mathbf{x}).
\end{equation}

\textit{Entropy and Mutual Information:}
\begin{equation}\label{eq:p4}
    H(\vb{x}) = H(\vb{x} \mid \vb{y}) + I(\vb{x};\vb{y}).
\end{equation}

\textit{Conditional Mutual Information and Conditional Entropy:}
\begin{equation}\label{eq:p5}
    I(\vb{x}; \vb{y} \mid \vb{z}) = H(\vb{x} \mid \vb{z}) - H(\vb{x} \mid \vb{y}\vb{z}).
\end{equation}

\citet{federici2020} assumes the following property when \(\vb{z_2}\) is a representation of \(\vb{x}\), which implies that the only source of stochastic in \(\vb{z_2}\) is from \(\vb{x}\):
\begin{equation}
    I(\vb{z_2}; \vb{y} \mid \vb{x}) = 0.
\end{equation}

\citet{federici2020} also proves the following proposition, which implies that \(\vb{z_2}\) must be as predictive as \(\vb{x}\) to predict \(\vb{y}\) when \(\vb{z_2}\) is a sufficient representation of \(\vb{x}\) for \(\vb{y}\).

\begin{proposition}[\cite{federici2020}]\label{proposition:sufficient-z1}
    Let \(\vb{x}\) and \(\vb{y}\) be random variables with joint distribution \(p(\vb{x}, \vb{y})\). Let \(\vb{z_2}\) is a representation of \(\vb{x}\). Then \(\vb{z_2}\) is sufficient for \(\vb{y}\) if and only if:
    \begin{equation}
        I(\vb{x}; \vb{y}) = I(\vb{y}; \vb{z_2}).
    \end{equation}
\end{proposition}

Similarly, we also assume stochasticity for \(\vb{z_1}\) is conditionally independent from any other variables in the system once \(\vb{x}\) and \(\vb{t}\) are observed:
\begin{equation} \label{eq:repr-z1}
    I(\vb{z_1}; \vb{y} \mid \vb{xt}) = 0.
\end{equation}

% \subsection{Proof for Proposition \ref{proposition:sufficient-condition}} \label{proof:sufficient-condition}

% \subsection{Proof for Corollary \ref{corollary:latent-contains-more}}\label{proof:latent-contains-more}

\subsection{Proof for Theorem \ref{theorem:z1-better}}\label{proof:z1-better}

First, we need to prove sufficient representation of inputs and side-channel information for a task implies that the representation does not discard any task-relevant information contained in the inputs and side channels.
\begin{proposition}\label{proposition:sufficient-condition}
    Let \(\vb{x}, \vb{t}, \vb{y}\) be random variables with joint distribution \(p(\vb{x, t, y})\). Let \(\vb{z_1}\) be a latent representation of \(\vb{x}\) with additional side-channel information \(\vb{t}\). Then \(\vb{z_1}\) is sufficient of \(\vb{x}\) and \(\vb{t}\) for \(\vb{y}\) if and only if \(I(\vb{xt}; \vb{y}) = I(\vb{y}; \vb{z_1})\).
\end{proposition}

\begin{proof}
    \begin{align*}
        I(\vb{xt}; \vb{y} \mid \vb{z_1}) 
        &\overset{(\ref{eq:p3})}{=} I(\vb{xt}; \vb{y}) - I(\vb{xt}; \vb{y}; \vb{z_1}) \\
        &\overset{(\ref{eq:p3})}{=} I(\vb{xt}; \vb{y}) - I(\vb{y}; \vb{z_1}) - I(\vb{y}; \vb{z_1} \mid \vb{xt}) \\
        &\overset{(\ref{eq:repr-z1})}{=} I(\vb{xt}; \vb{y}) - I(\vb{y}; \vb{z_1})
    \end{align*}
    Thus, \(\vb{z_1}\) is sufficient of \(\vb{x}\) and \(\vb{t}\) for \(\vb{y}\) (\(I(\vb{xt}; \vb{y} \mid \vb{z_1}) = 0\)) if and only if \(I(\vb{xt}; \vb{y}) = I(\vb{y}; \vb{z_1})\)
\end{proof}

Consequently, the mutual information between the sufficient representation \(\vb{z_1}\) and the label \(\vb{y}\) is at least as large as that of the input \(\vb{x}\) and the label \(\vb{y}\).
\begin{corollary}\label{corollary:latent-contains-more}
    Let \(\vb{x}, \vb{t}, \vb{y}\) be random variables with joint distribution \(p(\vb{x, t, y})\). Let \(\vb{z_1}\) be a sufficient latent representation of \(\vb{x}\) with additional side-channel information \(\vb{t}\) for \(\vb{y}\). Then, \(I(\vb{z_1}; \vb{y}) \geq I(\vb{x};\vb{y})\).
\end{corollary}

\begin{proof}
    \[I(\vb{y}; \vb{z_1}) = I(\vb{xt}; \vb{y}) \overset{(\ref{eq:p2})}{=} I(\vb{x}; \vb{y}) + I(\vb{t}; \vb{y} \mid \vb{x}) \overset{(\ref{eq:p1})}{\geq} I(\vb{x}; \vb{y}).\]
\end{proof}

The equality occurs when the side-channel information are irrelevant to the downstream task given the observation of the input. Lastly, we prove our main theorem.

\noindent\textbf{Theorem \ref{theorem:z1-better}.} \textit{
    Let \(\vb{x}, \vb{t}, \vb{y}\) be random variables with joint distribution \(p(\vb{x, t, y})\). Assume that \(\vb{z_1}\) is sufficient representation of \(\vb{x}\) with side-channel information \(\vb{t}\) for \(\vb{y}\), and \(\vb{z_2}\) is sufficient representation of \(\vb{x}\) for \(\vb{y}\). Then mutual information between \(\vb{z_1}\) and \(\vb{y}\) is at least as much as that between \(\vb{z_2}\) and \(\vb{y}\):
    \[
        I(\vb{z_1}; \vb{y}) \geq I(\vb{z_2}; \vb{y}).
    \]
}

\begin{proof}
    From Corollary \ref{corollary:latent-contains-more}, we have:
    \begin{align}
        I(\vb{z_1}; \vb{y}) &\geq I(\vb{x};\vb{y}) \\
        \overset{(\ref{eq:p4})}{\iff} -H(\vb{y} \mid \vb{z_1}) + H(\vb{y}) &\geq -H(\vb{y} \mid \vb{x}) + H(\vb{y})  \\
        \iff H(\vb{y} \mid \vb{z_1}) &\leq H(\vb{y} \mid \vb{x}) \label{eq:less-entropy}
    \end{align}

    Then we expand \(I(\vb{z_1}; \vb{y})\):
    \begin{align*}
        I(\vb{z_1}; \vb{y}) &\overset{(\ref{eq:p2})}{=} I(\vb{xz_1}; \vb{y}) - I(\vb{x}; \vb{y} \mid \vb{z_1}) \\
        &\overset{(\ref{eq:p2})}{=} I(\vb{x}; \vb{y}) + I(\vb{z_1}; \vb{y} \mid \vb{x})- I(\vb{x}; \vb{y} \mid \vb{z_1}) \\
        &\overset{(\ref{eq:p5})}{=} I(\vb{x}; \vb{y}) + \left( H(\vb{y} \mid \vb{x}) - H(\vb{y} \mid \vb{xz_1}) \right) - \left( H(\vb{y} \mid \vb{z_1}) - H(\vb{y} \mid \vb{xz_1}) \right) \\
        &= I(\vb{x}; \vb{y}) + H(\vb{y} \mid \vb{x}) - H(\vb{y} \mid \vb{z_1})
    \end{align*}

    Given Eq. \ref{eq:less-entropy} and Proposition \ref{proposition:sufficient-z1}, we have:
    \[
        I(\vb{z_1}; \vb{y}) \geq I(\vb{z_2}; \vb{y}) 
    \]
\end{proof}

\section{Formal Definition of Label Generation}\label{appendix:formal-def}

In Section \ref{subsec:label-generation}, we provided a compact notation for RDL labels, assuming that $k$-hop neighbors share the same entity type and that timestamps are discretized with natural integer indices. In this section, we provide the full notations for the label generation process. Formally, downstream labels can be represented as:
\begin{equation}\label{eq:yv_formal}
y_v^{(t_v,t_v+\Delta t]}
 = l\bigl({}^k_R\mathbf{X}_v^{(t_v,t_v+\Delta t]}\bigr),
\end{equation}

\begin{equation}\label{eq:Xv_formal}
{}^k_R\mathbf{X}_v^{(t_v,t_v+\Delta t]}
 = \bigl\{f_{\mathcal V}(u)\mid u\in\mathcal{N}_k^{(t_v,t_v+\Delta t]}(v)\land\varphi_{\mathcal V}(u)=R\bigr\},
\end{equation}

where \(l\) is a labeling set function and \({}^k_R\mathbf{X}_v^{(t_v, t_v + \Delta t]}\) denotes the set of \(k\)-hop neighbors' features of \(v\), of entity type \(R\), which appear within the prediction window, and \(\mathcal{N}_k^{(t_v, t_v + \Delta t]}\) denotes the set of temporal \(k\)-hop neighbors based on schema traversal. \(y_v^{(t_v,t_v+\Delta t]}\) corresponds to \(y_v^{(t)}\) in Eq.~\ref{eq:set-function}, which represents the label associated with the entity \(v\) at the cutoff timestamp \(t\).

\section{Experiments} \label{appendix:experiments}

In this section, we would like to provide further details regarding our experiments. All of the experiments are implemented in Pytorch \cite{paszke_pytorch_2019}, Pytorch Geometric \cite{fey_fast_2019}, and Pytorch Frame \cite{hu2024pytorch}. All of the experiments are executed on one of the following 4 machines: 1) AMD EPYC 7763 64-Core Processor and NVIDIA RTX A6000, 2) AMD EPYC 7543 32-Core Processor and NVIDIA RTX A5000, 3) AMD EPYC 7543 32-Core Processor and NVIDIA RTX A5000, and 4) AMD EPYC 7513 32-Core Processor and NVIDIA RTX A6000. Experiments are logged and kept tracked by Weights \& Biases \cite{wandb}. 
% Code is publicly available\footnote{https://anonymous.4open.science/r/tve-0BEE/}.

\subsection{Dataset and Task Descriptions}

Here, we provide details regarding datasets and tasks for our experiments. Datasets are proposed by RelBench \cite{robinson2024relbench}. In addition to the original tasks proposed by RelBench \cite{robinson2024relbench}, we also provide additional data limited tasks.

\paragraph{\texttt{rel-amazon}} 
The \texttt{rel-amazon} dataset is sourced from the Amazon e-commerce platform. It comprises four original downstream tasks \cite{robinson2024relbench}:
\begin{itemize}
      \item \texttt{user-churn}: predict whether a user remains inactive (i.e., has no interactions) over the next three months.
      \item \texttt{item-churn}: predict whether an item remains without interactions over the next three months.
      \item \texttt{user-ltv}: estimate the total monetary value of purchases made by a user over the next three months.
      \item \texttt{item-ltv}: estimate the total monetary value of purchases received by an item over the next three months.
  \end{itemize}

Additionally, we propose data limited tasks, which are conditioned on a certain attribute of entities up to the cutoff timestamp \(t\) and period \(\Delta t\), where \(\Delta t = 90\) days:
\begin{itemize}
    \item \texttt{item-churn-least-k-spending}: among \(k\) least bought products in the previous period \((t-\Delta t, t]\), predict products that remains without interactions over the next three months.
    \item \texttt{user-churn-top-k-spending}: among \(k\) top spending customers in the previous period \((t-\Delta t, t]\), predict those who will be inactive over the next three months.
    \item \texttt{item-ltv-least-k-spending}: among \(k\) least spending products in the previous period \((t-\Delta t, t]\), estimate the total monetary value of purchases received by an item over the next three months.
    \item \texttt{user-ltv-bad-reviews} (abbreviated as rel-amz/user-ltv-bad-rev. in Table \ref{tab:low-data-tasks}): among those customers who only left bad reviews (review score of 1, 2, or 3) in the previous period \((t-\Delta t, t]\), estimate the total monetary value of purchases made by those users over the next three months.
\end{itemize}

\paragraph{\texttt{rel-hm}} The \texttt{rel-hm} dataset is sourced from the H\&M relational database. Similar to \texttt{rel-amazon}, \texttt{rel-hm} comprises similar tasks as originally proposed by \cite{robinson2024relbench}:

\begin{itemize}
    \item \texttt{user-churn}: predict if a customer does not make any transactions in the next 7 days.
    \item \texttt{item-sales}: predict total sales for an article in the next 7 days.
\end{itemize}

Additionally, we also propose the following tasks, which are conditioned on a certain attribute of entities up to the cutoff timestamp \(t\) and period \(\Delta t\), where \(\Delta t = 7\) days:
\begin{itemize}
    \item \texttt{user-churn-top-k-spending}:  among \(k\) top spending customers in the previous period \((t-\Delta t, t]\), predict those who do not make any transactions in the next 7 days.
    \item \texttt{item-sales-top-k-spending}: among \(k\) top bought articles in the previous period \((t-\Delta t, t]\), estimate the total sales for those article in the next 7 days.
\end{itemize}

\subsection{Models}

In this section, we would like to provide further details regarding models included in this study.

\paragraph{Baseline} This is the base model adopted from RelBench \cite{robinson2024relbench} for node-level tasks.

\paragraph{Masked Autoencoder (MAE)} Building on GraphMAE \cite{hou2022graphmae}, we introduce the following modifications to adapt it to relational data. First, instead of masking entire node feature vectors, we apply cell-level masking—a technique well established in tabular learning \cite{yoon2020vime,bahri2022scarf,gorishniy2021revisitingdltabular}—and replace masked entries with values sampled from each column’s marginal distribution. This obviates the need for a dedicated mask token and a re-masking stage. Our decoder still retains a projection layer after the GNN encoder, a single-layer GNN decoder, and last linear heads. Second, because cell-level masking can alter a large fraction of each row, we reconstruct the full feature vector, as in \cite{yoon2020vime,liu2023flaky}. We limit reconstruction to numerical, categorical, and textual features: normalized numerical values and text embeddings are recovered using mean-squared error loss, while categorical features are recovered via cross-entropy loss. Any entries missing in the original database are excluded from the loss computation. The final training objective is the sum of these individual reconstruction losses for all tables.

\paragraph{Contrastive Learning (CTR)} Building on GraphCL \cite{you2020graph}, we make several adjustments similar to MAE. Specifically, we adopt the same masking augmentation as outlined in the previous paragraph. Also, follow SCARF \cite{bahri2022scarf}, we consider two views of the inputs to be original features and augmented features, and the final objective is to maximize agreements between latent representations of two views via a symmetric InfoNCE loss \cite{oord2018representation} for all tables.

\paragraph{Task Vector Estimation (TVE)} This is our proposed method, where we provide details in Section \ref{sec:predictive}. We also apply mask augmentation with probability (\(p = 0.15\)) similar to MAE and CTR for regularization. A noteworthy observation is that our method is root-table-centric, which means that the objective (Eq. \ref{eq:tve-loss}) is computed for a single table only. The hybrid models, namely MAE+TVE and CTR+TVE, also follow the same pipeline, with only changes in the decoder to cater for different loss objectives. For both hybrid models, we set the balancing coefficient \(\beta\) to \(0.1\) (see Eq. \ref{eq:combined-loss}).

\subsection{RelBench Experiments on Data Limited and Data Sufficient Tasks}

\begin{table}[t]
\caption{Performance across datasets and TVE variants in the sufficient-data regime \cite{robinson2024relbench}. Reported metrics are
ROC-AUC for classification tasks, and Mean Absolute Error for regression tasks. Split, Validation, and Test are abbreviated as S, V, and T for readability. Bold is the best performance.
% \czk{Why the baseline performance here is lower than one in relbench}
} 
\label{tab:supp-original-tasks}
\setlength{\tabcolsep}{2.15pt}
\resizebox{\textwidth}{!}{%
\begin{tabular}{@{}llcccccc@{}}
\toprule
\multirow{2}{*}{Task}               & \multirow{2}{*}{S} & \multicolumn{6}{c}{Model}                                                                                                                                  \\ \cmidrule(l){3-8} 
                                    &                    & TVE-1& TVE-2& MAE+TVE-1       & MAE+TVE-2                     & CTR+TVE-1                & CTR+TVE-2               \\ \midrule
\multicolumn{8}{c}{Classification} \\ \midrule
\multirow{2}{*}{\makecell[tl]{rel-amz/\\item-churn}} & V                  & 80.67 ± 0.06& 80.76 ± 0.08& 81.13 ± 0.04    & 81.19 ± 0.03              & 81.29 ± 0.09& 81.28 ± 0.10\\
                                    & T                  & 80.91 ± 0.13& 81.03 ± 0.10& 81.47 ± 0.04    & 81.49 ± 0.04              & 81.57 ± 0.13& 81.56 ± 0.12\\ \midrule
\multirow{2}{*}{\makecell[tl]{rel-amz/\\user-churn}} & V                  & 69.85 ± 0.03& 69.92 ± 0.04& 69.92 ± 0.04& 69.93 ± 0.05          & 70.04 ± 0.04& 70.03 ± 0.06\\
                                    & T                  & 69.67 ± 0.08& 69.78 ± 0.05& 69.78 ± 0.07& 69.80 ± 0.03          & 69.95 ± 0.02& 69.85 ± 0.07\\ \midrule
\multirow{2}{*}{\makecell[tl]{rel-hm/\\user-churn}}  & V                  & 70.35 ± 0.11& 70.38 ± 0.07& 70.29 ± 0.09& 70.35 ± 0.10          & 70.53 ± 0.04& 70.54 ± 0.16\\
                                    & T                  & 69.86 ± 0.17& 69.81 ± 0.17& 69.89 ± 0.09& 69.80 ± 0.11& 70.13 ± 0.22& 69.72 ± 0.51\\ \midrule
\multicolumn{8}{c}{Regression} \\ \midrule
\multirow{2}{*}{\makecell[tl]{rel-amz/\\item-ltv}}   & V                  & 48.183 ± 0.804& 47.667 ± 0.401& 45.590 ± 0.121& 45.573 ± 0.045        & 45.030 ± 0.100& 44.956 ± 0.078\\
                                    & T                  & 51.244 ± 0.229& 51.167 ± 0.341& 50.437 ± 0.395& 50.659 ± 0.417        & 50.012 ± 0.287& 49.753 ± 0.192\\ \midrule
\multirow{2}{*}{\makecell[tl]{rel-amz/\\user-ltv}}   & V                  & 12.442 ± 0.021& 12.455 ± 0.018& 12.429 ± 0.016& 12.437 ± 0.031       & 12.397 ± 0.010& 12.371 ± 0.018\\
                                    & T                  & 14.690 ± 0.058& 14.676 ± 0.034& 14.658 ± 0.028& 14.626 ± 0.019        & 14.622 ± 0.046& 14.577 ± 0.020\\ \midrule
\multirow{2}{*}{\makecell[tl]{rel-hm/\\item-sales}}  & V                  & 0.0637 ± 3e-4& 0.0636 ± 2e-4& 0.0627 ± 1e-4& 0.0624 ± 3e-4& 0.0628 ± 4e-4& 0.0627 ± 3e-4\\
                                    & T                  & 0.0554 ± 4e-4& 0.0553 ± 2e-4& 0.0544 ± 2e-4& 0.0543 ± 3e-4& 0.0547 ± 3e-4& 0.0543 ± 3e-4\\ \bottomrule
\end{tabular}%
}

\end{table}

In this section, we would like to provide further details regarding experiments reported in Section \ref{subsec:data-limited} and Section \ref{subsec:data-sufficient}. 

For pre-training, we train these models for \(81\) epochs with the max steps per epoch to be \(1000\) per GPU, which effectively means \(4000\) steps per epoch due to distributed training. Learning rate is \(0.001\) for the first \(50\) epochs and reduces to \(0.0001\) for the remaining epochs. During fine-tuning, we fine-tune the pre-trained models for \(20\) epochs (except to \texttt{rel-hm/item-sales-top200-spending} where we fine-tune the models for \(50\) epochs). We also perform light hyperparameter tuning where we vary the following hyperparameters: learning rate \(\in \{0.0075 , 0.001, 0.002, 0.005\}\) and number of channels \(\in \{128, 256\}\). For the original tasks proposed by RelBench, we adopt the learning rate of \(0.005\) and \(0.002\) for classification and regression tasks respectively. Similarly to the pre-training stage, we also perform distributed training on tasks with large amount of data. Data limited tasks are repeated \(10\) times (except to \texttt{rel-amazon/user-ltv-bad-reviews} and \texttt{rel-hm/item-sales-top200-spending} where we only repeat 5 times), while data sufficient tasks are repeated \(5\) times. All experiments use a batch size of 512 (reduced to 256 if batches exceed 24 GB VRAM), mean aggregation, two GNN layers, neighbor sampling of 128 then 64 hops, and uniform temporal sampling. Detailed hyperparameter differences are reported in Table \ref{tab:hyperparam-low} and Table \ref{tab:hyperparam-sufficient} for data limited and data sufficient tasks, respectively.

\begin{figure}[t]
    \centering
    \includegraphics[width=\textwidth]{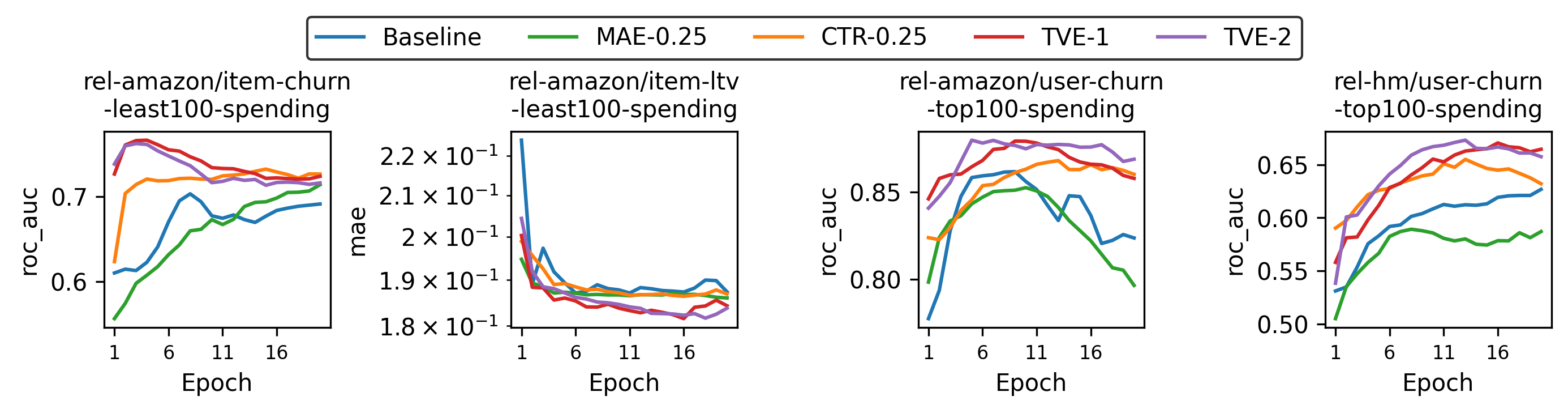}
    \caption{Test performance curves averaged over runs on data limited tasks.}
    \label{fig:main-curves}
\end{figure}

Fig. \ref{fig:main-curves} shows learning curves on low-data tasks, supplementing Table \ref{tab:low-data-tasks}. We can observe that TVE pre-training both accelerates convergence and outperforms MAE and CTR. Table \ref{tab:supp-original-tasks}, which complements Table \ref{tab:original-tasks}, compares all TVE variants in the data‐rich regime and shows that combining vanilla TVE with any standard SSL objective yields additional gains, underscoring their complementary strengths.

\begin{table}[t]
\caption{Hyperparameter settings for each model in Table \ref{tab:low-data-tasks}. Each configuration is in the format [learning rate, channel size].}
\label{tab:hyperparam-low}
\setlength{\tabcolsep}{2.15pt}
\resizebox{\textwidth}{!}{%
\begin{tabular}{@{}llccccccc@{}}
\toprule
\multirow{2}{*}{Task}      & \multicolumn{7}{c}{Model}                                                                                                                        \\ \cmidrule(l){3-9} 
\multicolumn{2}{l}{}                           & Baseline           & MAE-0.25           & MAE-0.5            & CTR-0.25           & CTR-0.5            & TVE-1              & TVE-2              \\ \midrule
\multirow{2}{*}{\makecell[tl]{rel-amz/\\item-churn}} & least50  & [0.001, 128]   & [0.005, 256]   & [0.001, 256]   & [0.002, 256]   & [0.005, 256]   & [0.001, 256]   & [0.001, 256]   \\ \cmidrule(l){2-9} 
                                    & least100 & [0.00075, 128] & [0.005, 128]   & [0.005, 128]   & [0.001, 256]   & [0.002, 256]   & [0.00075, 256] & [0.00075, 256] \\ \midrule
\multirow{2}{*}{\makecell[tl]{rel-amz/\\user-churn}} & top50    & [0.005, 128]   & [0.005, 256]   & [0.005, 256]   & [0.005, 256]   & [0.005, 128]   & [0.00075, 128] & [0.002, 256]   \\ \cmidrule(l){2-9} 
                                    & top100   & [0.002, 256]   & [0.002, 256]   & [0.005, 256]   & [0.002, 128]   & [0.001, 128]   & [0.005, 256]   & [0.005, 128]   \\ \midrule
\multirow{2}{*}{\makecell[tl]{rel-hm/\\user-churn}}  & top50    & [0.001, 128]   & [0.002, 128]   & [0.002, 256]   & [0.001, 256]   & [0.00075, 256] & [0.005, 128]   & [0.005, 128]   \\ \cmidrule(l){2-9} 
                                    & top100   & [0.005, 128]   & [0.002, 128]   & [0.001, 128]   & [0.00075, 256] & [0.001, 256]   & [0.005, 256]   & [0.002, 256]   \\ \midrule
rel-amz/item-ltv                 & least100 & [0.00075, 128] & [0.002, 128]   & [0.002, 256]   & [0.005, 256]   & [0.005, 256]   & [0.001, 256]   & [0.002, 256]   \\ \midrule
rel-amz/user-ltv                    & bad-rev. & [0.00075, 128] & [0.00075, 128] & [0.00075, 128] & [0.00075, 128] & [0.00075, 128] & [0.00075, 256] & [0.00075, 128] \\ \midrule
rel-hm/item-sales                   & top200   & [0.001, 256]   & [0.005, 256]   & [0.005, 128]   & [0.005, 256]   & [0.005, 256]   & [0.005, 256]   & [0.005, 256]   \\ \bottomrule
\end{tabular}%
}
\end{table}

\begin{table}[t]
\centering
\caption{Hyperparameter settings for each model in Table \ref{tab:original-tasks}. Each configuration is in the format [learning rate, channel size].}
\label{tab:hyperparam-sufficient}
\setlength{\tabcolsep}{2.15pt}
\resizebox{\textwidth}{!}{%
\begin{tabular}{@{}lccccccc@{}}
\toprule
\multirow{2}{*}{Task} & \multicolumn{7}{c}{Model}                                                                              \\ \cmidrule(l){2-8} 
                      & Baseline     & MAE-0.25     & MAE + TVE-1  & MAE + TVE-2  & CTR-0.25     & CTR + TVE-1  & CTR + TVE-2  \\ \midrule
rel-amz/item-churn    & [0.005, 128] & [0.005, 256] & [0.005, 128] & [0.005, 128] & [0.005, 256] & [0.005, 256] & [0.005, 256] \\ \midrule
rel-amz/user-churn    & [0.005, 128] & [0.005, 128] & [0.005, 128] & [0.005, 128] & [0.005, 128] & [0.005, 128] & [0.005, 256] \\ \midrule
rel-hm/user-churn     & [0.005, 256] & [0.005, 128] & [0.005, 256] & [0.005, 128] & [0.005, 256] & [0.005, 256] & [0.005, 256] \\ \midrule
rel-amz/item-ltv      & [0.002, 256] & [0.002, 256] & [0.002, 128] & [0.002, 128] & [0.002, 256] & [0.002, 256] & [0.002, 256] \\ \midrule
rel-amz/user-ltv      & [0.002, 128] & [0.002, 128] & [0.002, 256] & [0.002, 256] & [0.002, 256] & [0.002, 256] & [0.002, 256] \\ \midrule
rel-hm/item-sales     & [0.002, 256] & [0.002, 256] & [0.002, 256] & [0.002, 256] & [0.002, 256] & [0.002, 256] & [0.002, 256] \\ \bottomrule
\end{tabular}%
}
\end{table}

\subsection{Linear Probing Experiments}

\begin{table}[t]
\caption{Linear probing results. Reported metric is
ROC-AUC (higher is better) for classification tasks, and Mean Absolute Error (lower is better) for regression tasks. Split, Validation, and Test are abbreviated as S, V, and T for readability. Bold is the best performance.
% \czk{Why the baseline performance here is lower than one in relbench}
} 
\label{tab:linear-probing}
\setlength{\tabcolsep}{2.15pt}
\resizebox{\textwidth}{!}{%
\begin{tabular}{@{}llccccc@{}}
\toprule
\multirow{2}{*}{Task}               & \multirow{2}{*}{S} & \multicolumn{5}{c}{Model}                                                                                                                                  \\ \cmidrule(l){3-7} 
                                    &                    & Baseline       & MAE-0.25                 & CTR-0.25& TVE-1& TVE-2               \\ \midrule
\multicolumn{7}{c}{Classification} \\ \midrule
\multirow{2}{*}{\makecell[tl]{rel-amz/item-churn-least100-spending}}& V                  & 69.72 ± 4.98   & 61.18 ± 1.07    & 68.13 ± 1.78    & 81.57 ± 0.37    & \textbf{82.76 ± 0.43 }         \\
                                    & T                  & 63.01 ± 4.98   & 52.92 ± 0.79    & 56.96 ± 2.07    & \textbf{78.06 ± 0.65}    & 77.83 ± 0.95         \\ \midrule
\multirow{2}{*}{\makecell[tl]{rel-amz/user-churn-top100-spending}}& V                  & 84.53 ± 1.97   & 88.44 ± 0.29             & 88.62 ± 0.38& \textbf{91.42 ± 0.07}& 90.94 ± 0.14          \\
                                    & T                  & 78.36 ± 1.72   & 83.86 ± 1.04             & 81.63 ± 0.46& \textbf{86.56 ± 0.36}& 85.72 ± 0.38         \\ \midrule
\multirow{2}{*}{\makecell[tl]{rel-hm/user-churn-top100-spending}}& V                  & 69.43 ± 1.56& 64.27 ± 0.60             & \textbf{74.57 ± 0.73}& 71.34 ± 0.24& 70.09 ± 0.61          \\
                                    & T                  & 53.93 ± 1.46& 56.65 ± 0.53             & 64.22 ± 1.47& \textbf{65.33 ± 0.35}& 64.15 ± 0.61\\ \midrule
\multicolumn{7}{c}{Regression} \\ \midrule
\multirow{2}{*}{\makecell[tl]{rel-amz/item-ltv-least100-spending}}& V                  & 0.1753 ± 0.0005& 0.1751 ± 0.0001& 0.1755 ± 0.0005& 0.1749 ± 0.0011& \textbf{0.1746 ± 0.0013}\\
                                    & T                  & 0.1867 ± 0.0004& 0.1865 ± 0.0001& 0.1866 ± 0.0007& 0.1841 ± 0.0014& \textbf{0.1836 ± 0.0013}\\ \midrule
\multirow{2}{*}{\makecell[tl]{rel-hm/item-sales-top200-spending}}& V                  & 3.6312 ± 0.0118& 3.6948 ± 0.0126& 3.6524 ± 0.0123& 3.7160 ± 0.0140& \textbf{3.5519 ± 0.0056}\\
                                    & T                  & 3.5325 ± 0.0386& 3.4925 ± 0.0059& 3.5087 ± 0.1426& 3.4815 ± 0.0210& \textbf{3.4583 ± 0.0320}\\ \bottomrule
\end{tabular}%
}
\end{table}

We would like to give more results supplementing the one reported in Section \ref{subsec:linear-probing}. Table \ref{tab:linear-probing} reports numerical results that are visualized in Fig.~\ref{fig:linear-probing}. Specifically, in the linear probing setting, TVE consistently outperforms both MAE and CTR on every evaluated task. Notably, for \texttt{rel-amazon/item-churn-least100-spending} and \texttt{rel-amazon/item-ltv-least100-spending}, neither MAE-0.25 nor CTR-0.25 yields any improvement, highlighting their unreliability as pre-training objectives for downstream tasks.

\subsection{Experiments on Weakly Schema-dependent Tasks}
So far, we have only focused on scenarios where tasks are strictly defined by SQL queries. However, some tasks may weakly depend on the schema graph or relational entity graph---such as those based on manual labels or involving database errors. To quantify TVE’s sensitivity to the underlying relational entity graph's quality, we simulated these weak dependencies by running controlled ``noise'' experiments on the RelBench's \texttt{rel-amazon} database, using the following steps:
\begin{enumerate}
\item We randomly dropped 20\% and 40\% of the rows in the \texttt{review} table to corrupt the relational entity graph. The corrupted graph here is considered clean signals, and tasks depend on this newly created graph.
\item We treat the unaltered database as noisy database; thus, the downstream tasks defined in Step 1 are no longer fully dependent on this noisy database or its original graph structure. Therefore, a robust pre-training paradigm overfitting to the noisy relational entity graph cannot generalize well to the tasks in Step 1.
\item We compared three pre-training paradigms---MAE, CTR, and our TVE---on two tasks \texttt{user-churn-top100-spending} and \texttt{user-churn}. We then (a) pre-trained our models on the noisy graph (including task vector constructed from this noisy graph), and (b) fine-tuned (and evaluated) on the same noisy graph using labels generated from the clean database (constructed in Step 1). In other words, the only factor that deviates from prior experiments is the downstream labels, which are generated based on a simulated clean database.
\end{enumerate}

Table~\ref{tab:weakly-dependent} compares robustness to noises of different SSL paradigms. We can observe that even when the downstream tasks loosely depend on schema graph understanding, and when TVE is pre-trained on a noisy relational database, TVE’s task‐aware pre-training consistently outperforms standard SSL objectives. This demonstrates that our framework is robust to errors in the schema graph/relational entity graph and retains its effectiveness across a range of noise levels.

\begin{table}[t]
\centering
\caption{Performance across different model variants on the scenarios where tasks weakly depend on the relational entity graphs of \texttt{rel-amazon}. Reported metric is ROC-AUC.}
\label{tab:weakly-dependent}
\resizebox{\textwidth}{!}{%
\begin{tabular}{@{}lllcccc@{}}
\toprule
Task & &                                                                 & Baseline                         & MAE-0.25                         & CTR-0.25                         & TVE-1                            \\ \midrule
\multirow{4}{*}{user-churn-top100-spending} & \multirow{2}{*}{Linear Probing} & Drop 20\% & 75.42 ± 3.10                     & 82.26 ± 0.27                     & 78.98 ± 1.17                     & 86.79 ± 0.26                     \\ \cmidrule(l){3-7} 
                                            &                                 & Drop 40\% & 73.95 ± 1.76                     & 78.16 ± 0.46                     & 76.38 ± 1.80                     & 81.44 ± 0.17                     \\ \cmidrule(l){2-7} 
                                            & \multirow{2}{*}{Fine-tuning}    & Drop 20\% & 80.72 ± 1.14                     & 82.83 ± 0.41                     & 81.46 ± 2.31                     & 85.04 ± 0.92                     \\ \cmidrule(l){3-7} 
                                            &                                 & Drop 40\% & 76.39 ± 1.16                     & 76.97 ± 1.43                     & 75.72 ± 1.26                     & 78.29 ± 1.79                     \\ \midrule
\multirow{4}{*}{user-churn}                 & \multirow{2}{*}{Linear Probing} & Drop 20\% & 60.92 ± 0.61 & 65.62 ± 0.04 & 64.59 ± 0.01 & 70.17 ± 0.03 \\ \cmidrule(l){3-7} 
                                            &                                 & Drop 40\% & 61.95 ± 0.67 & 66.96 ± 0.03 & 66.14 ± 0.04 & 71.31 ± 0.01 \\ \cmidrule(l){2-7} 
                                            & \multirow{2}{*}{Fine-tuning}    & Drop 20\% & 70.09 ± 0.15 & 69.74 ± 0.21 & 69.98 ± 0.13 & 70.26 ± 0.10 \\ \cmidrule(l){3-7} 
                                            &                                 & Drop 40\% & 71.35 ± 0.06 & 70.61 ± 0.24 & 71.20 ± 0.17 & 71.37 ± 0.20 \\ \bottomrule
\end{tabular}%
}
\end{table}

\subsection{In-database Cross-entity Transfers}
Since our proposed pre-training approach is entity-type-centric, where the constructed task vector is limited to a certain entity type in the database, it is natural to question the transferability between pre-trained models on two different entity types in the same database. To better understand the separability of the pre-trained latent representation, we further conduct a linear probing experiment, where we investigate if an item-centric pre-trained model can be fine-tuned for a user-task, and vice versa. Table~\ref{tab:cross-entity} reports the results. TVE-1 (item) refers to the pre-trained model on the item's task vector, while TVE-1 (user) refers to the pre-trained model on the user's task vector. Surprisingly, even when the task vector is not matched with downstream tasks’ entity type, it can still yield better performance than other generic SSL methods. For example, TVE-1 (user) is only worse than TVE-1 (item) on \texttt{item-churn-least100-spending}, and TVE-1 (item) is as strong as MAE-0.25 on \texttt{user-churn-top100-spending}. Consequently, this experiment highlights the necessity to create an entity type-specific pre-training objective to maximize downstream performance gains.

\begin{table}[t]
\centering
\caption{Comparison of linear probing performance of TVE-1 pretrained on different entity types and other models. Reported metric is ROC-AUC.}
\label{tab:cross-entity}
\begin{tabular}{@{}lcc@{}}
\toprule
Model        & item-churn-least100-spending & user-churn-top100-spending \\ \midrule
Baseline     & 63.01 ± 4.98                 & 78.36 ± 1.72               \\
MAE-0.25     & 52.92 ± 0.79                 & 83.86 ± 1.04               \\
CTR-0.25     & 56.96 ± 2.07                 & 81.63 ± 0.46               \\
TVE-1 (item) & 78.06 ± 0.65                 & 83.28 ± 0.63               \\
TVE-1 (user) & 65.88 ± 2.25                 & 86.56 ± 0.36               \\ \bottomrule
\end{tabular}%

\end{table}

\subsection{Experiments on Cell-level Prediction}

While previous experiments focus on predictive modeling over the RelBench benchmark \cite{robinson2024relbench}, there exists other benchmarks such as 4DBInfer \cite{wang20244dbinfer} where tasks are defined differently. Specifically, RelBench focuses on inference an SQL-generated label attached to rows in the dimension table (the table that is static and does not contain timestamps), while 4DBInfer focuses on cell-level prediction.

We would like to show that our proposed framework is performant regardless of evaluation settings. We pre-trained TVE-1 on \texttt{seznam} dataset \cite{motl2015ctu}, and evaluated downstream performance on an entity attribute classification task called \texttt{charge}, where we would like to predict cells under column \texttt{sluzba} of entity table \texttt{dobito}. We reported 4 metrics---Accuracy, Macro F1, Micro F1, and Mean Reciprocal Rank (MRR)---on both validation and test splits. We leverage the \texttt{AutoCompleteTask} implementation provided by RelBench \cite{robinson2024relbench} to create the corresponding training table for cell-level prediction, and we split the dataset such that the cut-off timestamps for the validation set and test set are at 2015-03-31 and 2015-07-31, respectively. Table~\ref{tab:seznam} reports results for both linear probing and fine-tuning settings.

We can observe that the performance of MAE and CTR is not consistent across datasets, while TVE consistantly performs well and exhibits competitive performance (matching MAE for linear probing and matching CTR for full-finetuning). This is because both MAE and CTR have limitations. For example, CTR does not offer a separable representation after pre-training (low performance during linear probing), while MAE is not as performant as CTR during full fine-tuning. Moreover, these gains come despite a conceptual mismatch---TVE was optimized for next-state prediction on dimension tables, whereas the downstream tasks operate on fact-table rows. Regardless of evaluation setting, TVE consistently performs well across different evaluation settings.

\begin{table}[t]
\centering
\caption{Results on \texttt{seznam/charge}. Split, Validation, and Test are abbreviated as S, V, and T for readability.}
\label{tab:seznam}
\begin{tabular}{@{}llccccc@{}}
\toprule
Setting                         & Model                     & S & Accuracy     & Macro F1     & Micro F1     & MRR          \\ \midrule
\multirow{8}{*}{Linear Probing} & \multirow{2}{*}{Baseline} & V & 60.27 ± 1.22 & 28.89 ± 2.09 & 60.27 ± 1.22 & 76.96 ± 0.70 \\ \cmidrule(l){3-7} 
                                &                           & T & 59.36 ± 1.00 & 27.99 ± 1.86 & 59.36 ± 1.00 & 76.48 ± 0.50 \\ \cmidrule(l){2-7} 
                                & \multirow{2}{*}{MAE-0.25} & V & 71.03 ± 0.04 & 44.50 ± 1.24 & 71.03 ± 0.04 & 84.00 ± 0.03 \\ \cmidrule(l){3-7} 
                                &                           & T & 70.52 ± 0.05 & 43.32 ± 1.74 & 70.52 ± 0.05 & 83.76 ± 0.04 \\ \cmidrule(l){2-7} 
                                & \multirow{2}{*}{CTR-0.25} & V & 42.88 ± 0.09 & 12.74 ± 0.81 & 42.88 ± 0.09 & 64.56 ± 0.10 \\ \cmidrule(l){3-7} 
                                &                           & T & 41.11 ± 0.32 & 10.01 ± 0.48 & 41.11 ± 0.32 & 62.79 ± 0.21 \\ \cmidrule(l){2-7} 
                                & \multirow{2}{*}{TVE-1}    & V & 69.98 ± 0.21 & 44.20 ± 1.00 & 69.98 ± 0.21 & 83.44 ± 0.09 \\ \cmidrule(l){3-7} 
                                &                           & T & 69.90 ± 0.16 & 43.39 ± 0.64 & 69.90 ± 0.16 & 83.55 ± 0.08 \\ \midrule
\multirow{8}{*}{Fine-tuning}    & \multirow{2}{*}{Baseline} & V & 81.13 ± 0.21 & 62.02 ± 0.76 & 81.13 ± 0.21 & 89.96 ± 0.12 \\ \cmidrule(l){3-7} 
                                &                           & T & 80.97 ± 0.50 & 63.43 ± 0.83 & 80.97 ± 0.50 & 89.91 ± 0.28 \\ \cmidrule(l){2-7} 
                                & \multirow{2}{*}{MAE-0.25} & V & 81.18 ± 0.15 & 62.00 ± 2.37 & 81.18 ± 0.15 & 89.99 ± 0.09 \\ \cmidrule(l){3-7} 
                                &                           & T & 81.08 ± 0.32 & 63.31 ± 1.00 & 81.08 ± 0.32 & 89.97 ± 0.18 \\ \cmidrule(l){2-7} 
                                & \multirow{2}{*}{CTR-0.25} & V & 81.21 ± 0.17 & 63.28 ± 1.05 & 81.21 ± 0.17 & 90.01 ± 0.10 \\ \cmidrule(l){3-7} 
                                &                           & T & 81.31 ± 0.41 & 63.69 ± 0.92 & 81.31 ± 0.41 & 90.10 ± 0.22 \\ \cmidrule(l){2-7} 
                                & \multirow{2}{*}{TVE-1}    & V & 81.39 ± 0.22 & 63.92 ± 1.17 & 81.39 ± 0.22 & 90.11 ± 0.12 \\ \cmidrule(l){3-7} 
                                &                           & T & 81.26 ± 0.29 & 63.82 ± 0.75 & 81.26 ± 0.29 & 90.07 ± 0.16 \\ \bottomrule
\end{tabular}
\end{table}

\subsection{Performance Sensitivity Across Splits and Tasks}

\begin{figure}[t]
    \centering
    \includegraphics[width=\linewidth]{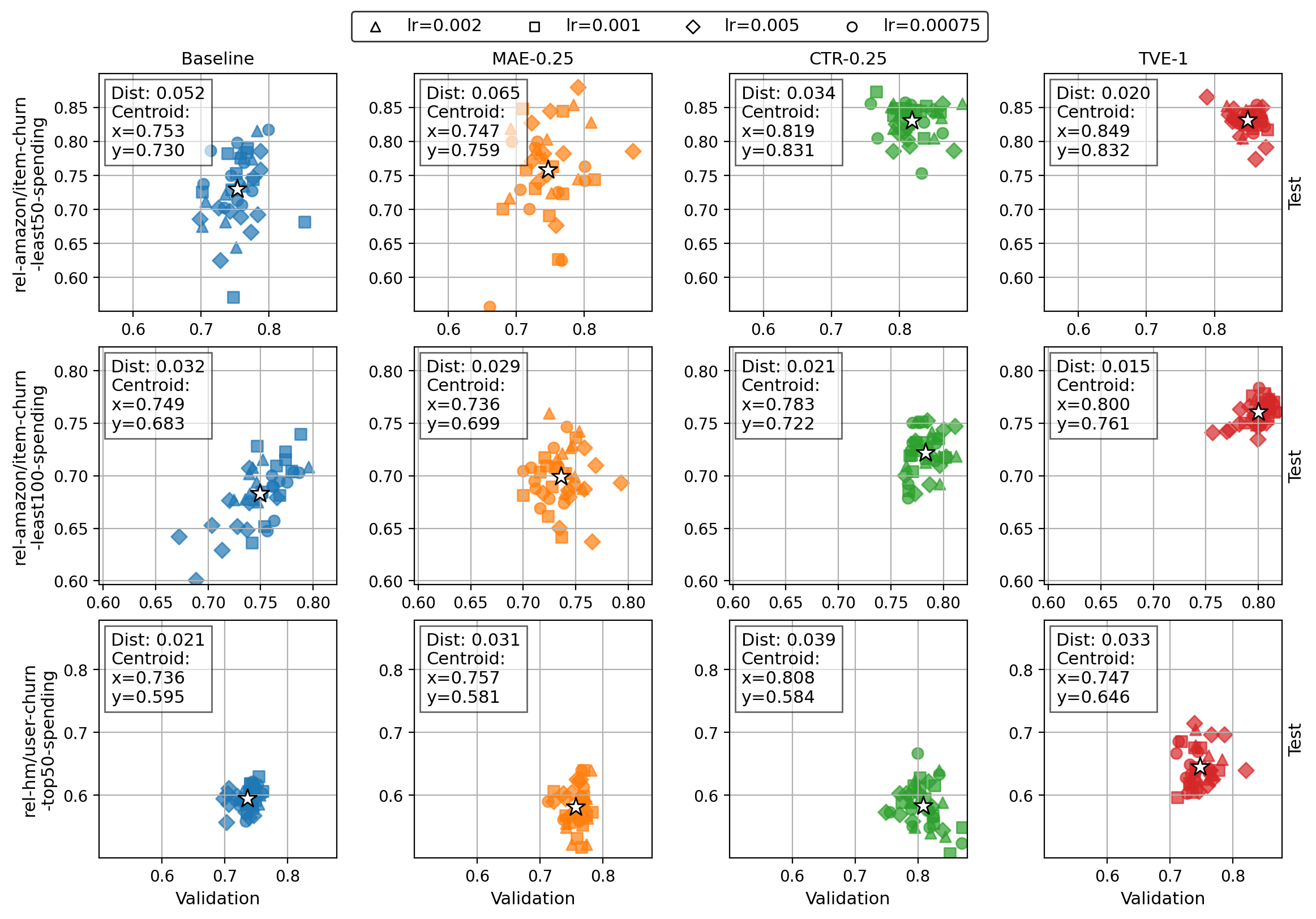}
    \caption{Additional Results for Validation vs. Test scatter plots for low-data node classification tasks across varying learning rates. Distances to centroid (denoted \ding{73}) are used to measure performance variance.}
    \label{fig:flaky-scatter-v2}
\end{figure}

To further assess model robustness, Fig.~\ref{fig:flaky-scatter-v2} presents additional scatterplots for three more tasks, namely \texttt{rel-amazon/item-churn-least50-spending}, \texttt{rel-amazon/item-churn-least100-spending}, and \texttt{rel-hm/user-churn-top50-spending}, with the purpose of complementing the results reported in Fig. \ref{fig:flaky-scatter}. Across all three tasks, TVE achieves higher validation and test scores with consistently smaller average distances to the centroid, whereas traditional SSL methods suffer from recency bias, where they exhibit inflated validation performance but degraded test results (e.g.\ \texttt{rel-hm/user-churn-top50-spending}). These findings underscore TVE’s stability to hyperparameter choices and the last linear head's weight initialization.

\subsection{Complexity Analysis}

In this section, we present complexity analysis on task vector construction and provide comparison of pre-training runtimes between our proposed approach and the baseline SSL methods.

\subsubsection{Task Vector Construction}

First, we provide formal worst-case time complexity of task vector construction, and then provide empirical analysis of task vector construction on \texttt{rel-amazon} and \texttt{rel-hm} via elapsed time and peak memory usages.

\begin{remark}
    Let the number of rows per table be \(\mathcal{O}(N)\), the number of timestamps be \(T\), the branching factor of relational entity graph (the input graph for learning) be \(b\), the number of aggregation functions be \(r\), the branching factor of schema graph (the blueprint of the Relational Database) be \(p\), and the number of hops for generating the task vector be \(k\). Then, the time complexity of task vector construction is 
    \[T_{\text{TVE}} = \mathcal{O}(rNTp^kb^k).\]
\end{remark}

\begin{proof}
    We would like to compute the computational cost at each step for constructing the task vector. First, we compute the cost taken for a single path from the schema graph perspective (Step 1 - 3), and then sum up to get the final cost (Step 4). Remember that for every schema graph path, there are many paths fall under the same schema graph path since schema graph is just a blueprint of the input graph.

    \textbf{Step 1.~} Joining with the time table, which is a necessary step regardless of pre-training paradigm, costs $\mathcal{O}(NT)$.
    
    \textbf{Step 2.~} Next, every join causes each row to fan out by a factor of $\mathcal{O}(b)$, so the cost of joining up to $u$ successive times is:
        
        $$
        \mathcal{O}(NTb + NTb^2 + ... + NTb^u) = \mathcal{O}(NT\sum_{i=1}^u b^i).
        $$
        
        For $b > 1$, the geometric sum is scaled on the order of $b^u$:
        
        $$
        \sum_{i=1}^u b^i = \frac{b^{u+1} - b}{b-1} \approx \frac{b^{u+1}}{b-1} = \frac{b}{b-1}b^{u} = \Theta(b^u).
        $$
        
        Therefore, the total cost of for joining up to length $u$ is $\mathcal{O}(NTb^u)$.
        
    \textbf{Step 3.~} One aggregation of $r$ functions over the joined rows costs $\mathcal{O}(rNTb^u)$, so every path (from the schema graph perspective) cost $\mathcal{O}(NTb^u + rNTb^u) = \mathcal{O}(rNTb^u)$.
    
    \textbf{Step 4.~} There are $\mathcal{O}(p^u)$ paths of length $u$ from the schema graph perspective, summing the cost:
    
    $$
    T_{\text{TVE}} = \sum_{u=1}^k \left[p^u \times \mathcal{O}(rNTb^u)\right] = \mathcal{O}\left(rNT\sum_{u=1}^k (pb)^u\right) = \mathcal{O}(rNTp^kb^k)
    $$
\end{proof}

Therefore, the cost blows up exponentially in $k$ through both $p^k$ and $b^k$, with linear dependence on $r$, $N$, and $T$. In practice, branching factor of the schema graph $p$ is typically small, and the schema graph is small, leading to small $k$. Furthermore, a path is only meaningful if it contains at least one fact table (the table containing timestamps), so that we can construct the predictive pre-training objective. Finally, this operation is only done once, and is not involved during pre-training optimization.

To demonstrate our point, we would like to report the time and peak memory usage for constructing task vector across datasets in Table~\ref{tab:time-complexity-tve}. As we can see from the table, this operation is feasible even with large-scale databases, where \texttt{rel-amazon} and \texttt{rel-hm} contain approximately 24M and 33M rows, respectively. For our experiments, the maximum hop is 2, while the possible aggregation functions include \texttt{MEAN}, \texttt{MIN}, \texttt{MAX}, \texttt{SUM}, \texttt{COUNT}, and \texttt{STDDEV}  for numerical values, and \texttt{MODE}, \texttt{COUNT}, and \texttt{COUNT DISTINCT} for categorical values. In industry scenario, such cost is expendable.

\subsubsection{Pre-training Runtimes}
We present an empirical comparison of per‐batch pre-training runtimes (batch size = 512) for all SSL methods. Using a shared RDL backbone, we measure the wall‐clock time for a single training step, averaged over 100 runs per model. As shown in Table \ref{tab:time-complexity}, TVE achieves the lowest overhead, since it computes losses to approximate task vectors only for root‐table entities. In contrast, MAE and CTR must compute losses for all nodes in the sampled subgraphs, and CTR incurs additional cost from its double forward passes and pairwise distance calculations.

\begin{table}[t]
\centering
\caption{Elapsed time and peak memory usage for different datasets given root tables.}
\label{tab:time-complexity-tve}
\resizebox{\textwidth}{!}{%
\begin{tabular}{@{}lllll@{}}
\toprule
Model & \multicolumn{1}{c}{rel-amazon/item} & \multicolumn{1}{c}{rel-amazon/user} & \multicolumn{1}{c}{rel-hm/item} & \multicolumn{1}{c}{rel-hm/user} \\ \midrule
TVE-1 & 64.59 seconds / 39.98 GB            & 90.17 seconds / 40.07 GB            & 19.38 seconds / 4.62 GB         & 11.74 seconds / 19.85 GB        \\
TVE-2 & 382.75 seconds / 40.03 GB           & 652.85 seconds / 52.67 GB           & 98.03 seconds / 16.15 GB        & 871.26 seconds / 148.19 GB      \\ \bottomrule
\end{tabular}%
}
\end{table}

\begin{table}[t]
\caption{Time complexity of different SSL methods for every step in seconds.}
\label{tab:time-complexity}
\resizebox{\textwidth}{!}{%
\begin{tabular}{@{}llcccc@{}}
\toprule
\multirow{2}{*}{Dataset}    & \multirow{2}{*}{Root Table}& \multicolumn{4}{c}{Model}                                             \\ \cmidrule(l){3-6} 
                            &                               & MAE             & CTR             & TVE-1           & TVE-2           \\ \midrule
\multirow{2}{*}{rel-amazon} & user                          & 0.0554 ± 0.0071& 0.0587 ± 0.0088& 0.0445 ± 0.0102& 0.0432 ± 0.0165\\
                            & item                          & 0.0583 ± 0.0065& 0.0663 ± 0.0059& 0.0440 ± 0.0089& 0.0478 ± 0.0101\\ \midrule
\multirow{2}{*}{rel-hm}     & user                          & 0.0610 ± 0.0115& 0.0615 ± 0.0062& 0.0504 ± 0.0288& 0.0466 ± 0.0053\\
                            & item                          & 0.0814 ± 0.0075& 0.3110 ± 0.0552& 0.0609 ± 0.0166& 0.0571 ± 0.0038\\ \bottomrule
\end{tabular}%
}
\end{table}

\section{Further Discussions} \label{app:further}
\subsection{On Designing Better Approximation of Sufficient Statistics}

Designing richer approximations of sufficient statistics is indeed a vital and challenging direction. For example, beyond fixed aggregators like task vectors, we can employ a small neural “set encoder” (e.g., DeepSets \cite{zaheer2017deepsets} or transformer-based pooling) that learns to compress each next-window set into a vector. This approach can, in principle, capture all information in the set without hand-crafting statistics. However, training two networks---one to encode the set and one to predict the label---introduces instability. For instance, a GAN-style formulation (where the encoder acts as a generator and the predictor as a discriminator) risks mode collapse, mapping diverse inputs to the same latent code. Therefore, designing a sophisticated loss function to avoid mode collapse is a challenge for this strategy.

Our current task-vector objective optimizes only one encoder for input graph, which makes training stable and efficient. Despite being simpler, these vectors capture key moments of the distribution and are not susceptible to GAN-style collapse. They provide a reliable surrogate that practitioners can use immediately, while leaving more complex, learnable summaries to future work.

\subsection{Sufficiency of Simple Statistics}

Computing minimal sufficient representations has long been a grand challenge. Because direct evaluation of mutual information is often intractable, \citet{alemi2017deep} instead optimize a tractable lower bound, while \citet{belghazi2018mutual} turn to neural estimators to approximate mutual information directly.

Given the above challenges of sufficient statistics approximation, we would like to provide complementary arguments showing why---and under what conditions---simple statistics can suffice, and how we mitigate the information loss when compared with sufficient statistics in practice.

\textbf{Statistical sufficiency in common distributions.~} In many distributions, a small collection of aggregators is provably sufficient. Notable examples include:

\begin{itemize}
    \item The sum of samples is sufficient for exponential, Poisson, and Bernoulli distributions.
    \item The max is sufficient for a uniform distribution over a bounded interval.
    \item The mean is sufficient for a normal distribution with known variance.
\end{itemize}

Whenever the underlying dynamics match one of these canonical families, no information is lost by reducing the full set of observations to the corresponding statistic.

\textbf{Enriching the task vector to minimize loss.~} Real-world label‐derivation queries often depend on different columns, thus leading to potential different distributions. To guard against this, we augment our task vectors with as many simple statistics---include \texttt{MEAN}, \texttt{MIN}, \texttt{MAX}, \texttt{SUM}, \texttt{COUNT}, and \texttt{STDDEV} for numerical values, and \texttt{MODE}, \texttt{COUNT}, and \texttt{COUNT DISTINCT} for categorical values. This ``overcomplete'' summary ensures that, even if some statistics are redundant, we capture a broader range of moment and order information, thereby shrinking the worst‐case information gap.

\textbf{Set‐theoretic recovery and downstream expressivity.~} From a set‐theoretic standpoint, certain combinations of simple statistics can perfectly reconstruct small multisets. For instance, if exactly three purchases occur in the prediction window, then the triplet \(\{\texttt{MIN}, \texttt{MAX}, \texttt{SUM}\}\) uniquely determines the multiset. More generally, accurate pre‐training on these statistics helps the model learn an implicit approximation of the original value‐set, which in turn improves its ability to approximate any downstream labeling function---since most such functions operate on those same sets (cf. Eq.~\ref{eq:set-function}).

\textbf{Empirical validation in tabular domains.~} Prior work on Deep Feature Synthesis (DFS) \cite{kanter2015deep} demonstrates that stacking joins with simple aggregations often matches more expressive graph‐based architectures on many benchmarks \cite{wang20244dbinfer, wang2025griffin}. While DFS features incur theoretical information loss relative to a full relational graph encoding, their competitive accuracy shows that richly engineered statistics can suffice in practice. Our investigation is orthogonal: we systematically characterize all node-level SQL-generated labels as set functions and show how a task‐vector approach can act as surrogate representation for next-window sets. Additionally, our task vector includes similar aggregators to that of DFS.

Together, these points explain why simple statistics are not only theoretically grounded but also practically powerful---and how we effectively minimize any information loss.

